\documentclass[sn-mathphys-ay]{sn-jnl}% Math and Physical Sciences Author Year Reference Style
\usepackage{graphicx}%
\usepackage{multirow}%
\usepackage{amsmath,amssymb,amsfonts}%
\usepackage{amsthm}%
\usepackage{mathrsfs}%
\usepackage[title]{appendix}%
\usepackage{xcolor}%
\usepackage{textcomp}%
\usepackage{manyfoot}%
\usepackage{booktabs}%
\usepackage{algorithm}%
\usepackage{algorithmicx}%
\usepackage{algpseudocode}%
\usepackage{listings}%

\usepackage{changepage}
\usepackage{adjustbox}   
\usepackage{natbib}
\usepackage[right]{lineno}
\usepackage{algorithm}%
\usepackage{algorithmicx}%
\usepackage{algpseudocode}%

\usepackage{float} % Required for the H placement specifier

\begin{document}

\title[Article Title]{BicliqueEncoder: An Efficient Method for Link Prediction in Bipartite Networks using Formal Concept Analysis and Transformer Encoder}


\author*[1]{\fnm{Hongyuan} \sur{Yang}}\email{yanghongyuan27@gmail.com}
\equalcont{These authors contributed equally to this work.}

\author[1]{\fnm{Siqi} \sur{Peng}}\email{peng.siqi@iip.ist.i.kyoto-u.ac.jp}
\equalcont{These authors contributed equally to this work.}

\author[1]{\fnm{Akihiro} \sur{Yamamoto}}\email{yamamoto.akihiro.5m@kyoto-u.ac.jp}

\affil*[1]{\orgdiv{Graduate School of Informatics}, \orgname{Kyoto University}, \orgaddress{\street{Sakyo Ward}, \city{Kyoto City}, \postcode{606–8501}, \state{Kyoto}, \country{JAPAN}}}

% \affil[2]{\orgdiv{Department}, \orgname{Organization}, \orgaddress{\street{Street}, \city{City}, \postcode{10587}, \state{State}, \country{Country}}}

% \affil[3]{\orgdiv{Department}, \orgname{Organization}, \orgaddress{\street{Street}, \city{City}, \postcode{610101}, \state{State}, \country{Country}}}


\abstract{We propose a novel and efficient method for link prediction in bipartite networks, using \textit{formal concept analysis} (FCA) and the Transformer encoder. Link prediction in bipartite networks finds practical applications in various domains such as product recommendation in online sales, and prediction of chemical-disease interaction in medical science. Since for link prediction, the topological structure of a network contains valuable information, many approaches focus on extracting structural features and then utilizing them for link prediction. Bi-cliques, as a type of structural feature of bipartite graphs, can be utilized for link prediction. Although several link prediction methods utilizing bi-cliques have been proposed and perform well in rather small datasets, all of them face challenges with scalability when dealing with large datasets since they demand substantial computational resources. This limits the practical utility of these approaches in real-world applications. To overcome the limitation, we introduce a novel approach employing iceberg concept lattices and the Transformer encoder. Our method requires fewer computational resources, making it suitable for large-scale datasets while maintaining high prediction performance. We conduct experiments on five large real-world datasets that exceed the capacity of previous bi-clique-based approaches to demonstrate the efficacy of our method. Additionally, we perform supplementary experiments on five small datasets to compare with the previous bi-clique-based methods for bipartite link prediction and demonstrate that our method is more efficient than the previous ones.}




\maketitle

\section{Introduction}\label{sec1}

\textit{Link prediction} in~\textit{bipartite networks}, or \textit{bipartite link prediction}, is an important task for finding potential relations between two groups of entities. A bipartite network comprises two disjoint sets of entities and a set of edges, each connecting only two nodes from different sets. Two entities connected with an edge are considered related~(\cite{bipartite3,bipartite4,nouse}). Real-world bipartite networks often have missing or unobserved links, giving rise to the link prediction task for predicting the absence or presence of these links~(\cite{linkprediction1,linkprediction2,linkprediction3}). Bipartite link prediction task has practical applications such as product recommendation in e-commerce (e.g., recommending products to customers based on co-purchases)~(\cite{movie1,commonneighbors}), chemical-disease interactions prediction in medical science to identify potential treatments~(\cite{chemical-disease1,chemical-disease2, chemical-disease3,proteinexample}) and friend suggestions in social networking services (e.g., suggesting mutual friends on platforms like Facebook)~(\cite{bipartite4,bipartite3,linkprediction3}). 

Most bipartite link prediction methods rely on structural features of the networks~(\cite{weakclique}). Among all structural features, recent research highlights the importance of \textit{bi-cliques} on bipartite link prediction~(\cite{weakclique,missbin,linkbipartite,matrixnega}). A \textit{bi-clique} is a complete sub-graph of a bipartite network where each entity in one set is connected to each entity in the other set. Bi-cliques can be regarded as clusters of strongly interconnected entities, representing cohesive groups within networks. For example, in a chemical-disease network, a bi-clique represents a group of similar chemicals and their affected diseases. Previous research has proposed the \textit{structural hole theory}~(\cite{structurehole2}), suggesting that two strongly connected clusters sharing numerous nodes likely belong to a larger cluster, implying that current unlinked pairs within these clusters are likely to be connected. Thus, link prediction can be performed by extracting bi-cliques and analyzing their relationships. Various methods utilizing bi-cliques have been proposed and shown good performance in small datasets.

However, all them face a severe problem: they are inapplicable to large-scale datasets, as they need to extract \textbf{all} bi-cliques from the network, which requires significant computational resources. Such an exhaustive enumeration is usually conducted by converting the network into an equivalent \textit{formal context} and performing a full process of \textit{formal concept analysis} (FCA) to extract all \textit{formal concepts}, each corresponding to a maximal bi-clique in the original bipartite network, from the converted formal context. The time complexity of the full process of FCA is $O(C)$ where $C$ is the number of formal concepts, which is exponential to the number of entities and links in the network. Therefore, this bi-cliques extraction step may take a long time. Furthermore, after extraction, some methods~(\cite{boa,bert4fca}) require training complex neural-network models from \textbf{all} extracted bi-cliques, consuming extensive RAM and GPU memory. Consequently, these methods struggle with large and dense bipartite networks due to their high computational cost.

To overcome this problem, we propose a novel and efficient method called \textit{BicliqueEncoder}, which can be applied to large datasets with manageable and adjustable memory usage and execution time. The core feature of this method is that it does not extract all bi-cliques, but only extracts those that are considered ``significant'' in terms of their size. As introduced above, the process of extracting bi-cliques is equivalent to that of extracting formal concepts from a formal context converted from the original bipartite network. In the field of FCA, researchers have proposed the idea of \textit{iceberg concept lattice}, which helps identify the possibility of only extracting the formal concepts of large sizes~(\cite{iceberglattice}) without needing to exhaustively enumerate all formal concepts. Since large formal concepts correspond to large bi-cliques which contain more information on the relations between entities, we consider they might be more significant and contribute more to link prediction than smaller bi-cliques. Hence, if we only extract and process the bi-cliques corresponding to the formal concepts in the iceberg concept lattice, we may be able to capture the key information provided by the bi-cliques with a lot fewer computational resources.

We adopt the \textit{Transformer encoder}~(\cite{Transformer}) to learn useful information and make link predictions because we believe that it can counterbalance the information loss caused by using only the extracted subset of all bi-cliques. Certainly, by applying the strategy that only extracts ``significant'' bi-cliques, we will leave out the information of all other ``insignificant'' bi-cliques in the bipartite network, which may cause a negative influence on the final prediction performance. For example, some nodes may not be included in the extracted bi-cliques. In such a case, the model cannot make predictions about such nodes when trained only on the extracted bi-cliques. To counterbalance such negative influence, we use the Transformer encoder to effectively process and capture the information from the original bipartite network as well as the extracted significant bi-cliques, and use the captured information to make link predictions. Transformer is a sequence-to-sequence deep learning framework adept at learning the dependencies within sequential data. Since bi-cliques are equivalent to formal concepts, and a formal concept can be treated as an unordered sequence (\textit{i.e.} a set) we consider the Transformer to be proper for processing and capturing the information from bi-cliques. We have conducted experiments on several real-world datasets to show that after adjusting the architecture of the Transformer encoder, it can indeed capture pertinent information during training and leverage it for good link prediction performance. 

Our main contributions are summarized as follows:
\begin{itemize}
    \item We introduce BicliqueEncoder, a novel method for bipartite link prediction using the Transformer encoder architecture. It can be applied to large datasets that the previous bi-cliques-based methods cannot handle, and its requirement of computational resources is flexibly adjustable. We conduct experiments on five large datasets that exceed the capacity of traditional bi-clique-based approaches and five small datasets to demonstrate the efficacy and efficiency of our method.
%    \item We propose an approach to address the limitation of FCA-based methods being inapplicable to large datasets by leveraging iceberg concept lattices. This strategy provides a solution applicable not only to our ConceptEncoder but also potentially to other FCA-based methods. 
    \item We demonstrate the contribution of the information provided by the extracted significant bi-cliques to prediction performance in ablation experiments. Moreover, we show that our proposed networks also can directly extract the information from the original bipartite network. It can achieve good performance even without the information from bi-cliques. 
    % \item We propose a novel graph Transformer~(\cite{graphtransformer} architecture which is designed specifically for bipartite link prediction. It can effectively learn the relationship between pairs of nodes from different sets in a bipartite network and achieve good performance on bipartite link prediction. 
\end{itemize}

The remaining part of this paper is organized as follows. In Section~\ref{sec2}, we start with some preliminaries, including bipartite networks, bi-cliques, FCA, and the Transformer. In Section~\ref{sec3}, we introduce some related link prediction methods utilizing bi-cliques. In Section~\ref{sec4}, we introduce and analyze our method BicliqueEncoder. In Section~\ref{sec5}, we describe our experiments and datasets and discuss the results. Finally, in Section~\ref{sec6}, we draw conclusions and discuss the limitations and our future work. 

\section{Preliminaries}\label{sec2}

\subsection{Bipartite Networks and Bi-cliques}\label{subsec1}

\begin{figure}[!htbp]
    \centering
    \includegraphics[width=\textwidth]{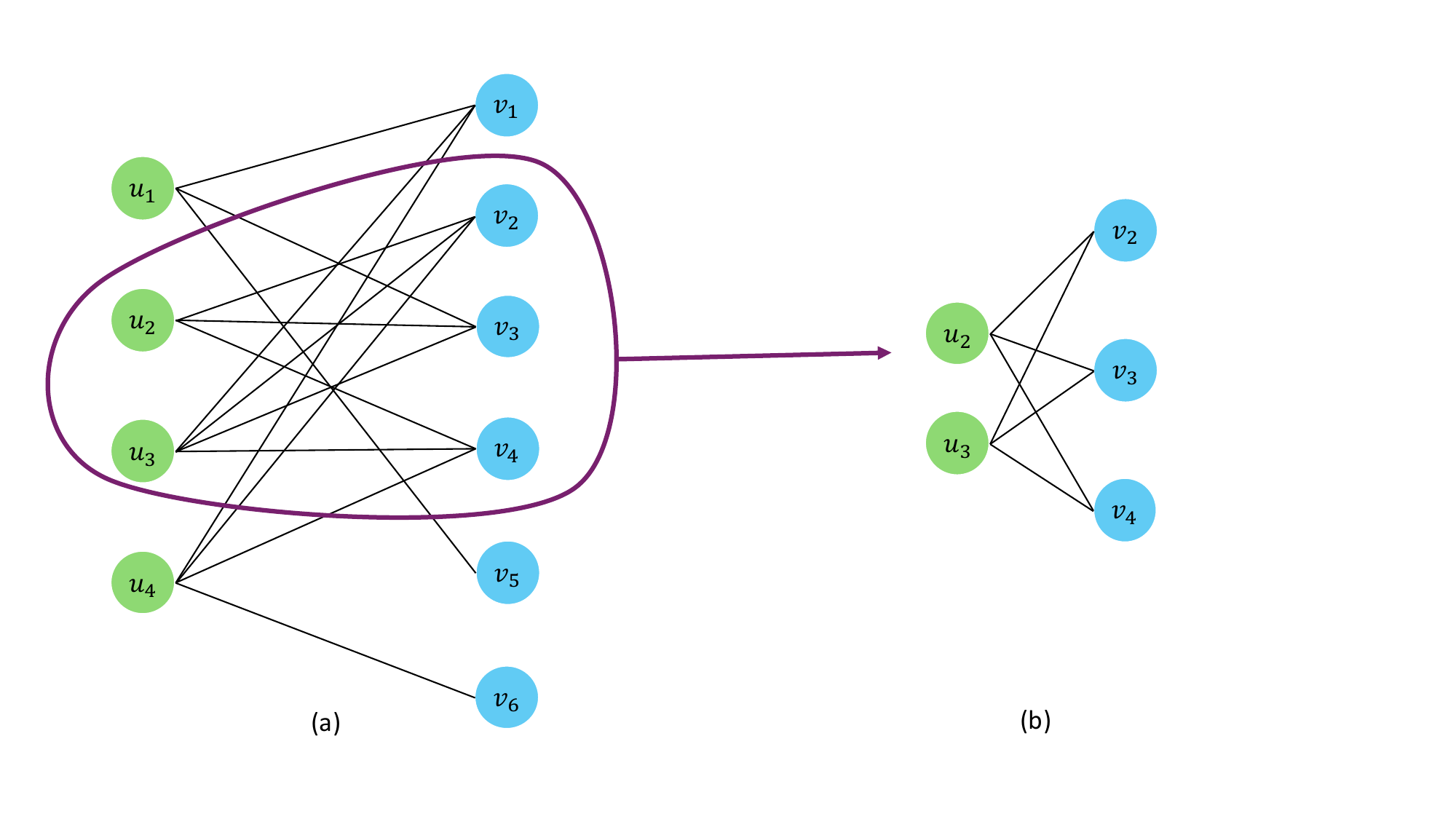}
    \caption{An example of a bipartite network(also refered as bipartite graph) $(U,V,E)$ and two of its bi-cliques. The nodes in lime form the node set $U$, and the nodes in light blue form the node set $V$. The sub-graph framed in purple is a maximal bi-cliques of the network. }
    \label{bipartitenetwork}
\end{figure}

{\bf Bipartite Networks:} A {\em bipartite network\/} $C$  is usually formalized  as a {\em bipartite graph\/}, a triple $(U, V, E)$,  where $U$ and $V$ are two disjoint sets of nodes and $E\subseteq U \times V$  is a set of edges.
An edge connecting a node $u \in U$ to another node $v \in V$ is denoted by $(u, v)$. 
Figure~\ref{bipartitenetwork}(a) illustrates an example of a bipartite graph $(U, V, E)$.  
In the graph all elements in $U$ are laid on the left-hand side and colored lime, and 
all elements in $V$ are laid on the right-hand side and colored blue. 
Note that every edges connects a node in $U$ and a node in $V$. 

{\bf Subgraphs, Bi-cliques and Maximal Bi-cliques:} 
A {\em subgraph\/} of a bipartite graph $G=(U, V, E)$ is a bipartite graph $G_1(U_1, V_1, E)$ such that
$U_1 \subseteq U$, $V_1 \subseteq V$,  $E_1\subseteq E\cap (U_1 \times V_1)$.
The  subgraph $G_1$ of  is a {\em bi-clique\/} of $G$ if $E_1 =U_1 \times V_1$. 
We say that the bi-clique  $G_1$ is {\em maximal\/} if   $G_1$ is not a subgraph of 
any other maximal bi-clique of  $G$,  that is, for all $G_2 = (U_2, V_2, E_2)$  such that
$U_1 \subseteq U_2 \subseteq U$ and $V_1 \subseteq V_2 \subseteq V$,
$E_2 = U_2 \times  V_2$ holds if and only if $U_1 = U_2$ and $V_1 = V_2$. 
Figure~\ref{bipartitenetwork}(b) gives an example of a maximal bi-clique.

% \textbf{Bipartite Networks}: A \textit{bipartite network} $C$ (also referred to as bipartite graph) is a triple $(U, V, E)$ where $U$, $V$ are two disjoint sets of \textit{nodes} and $E$ is a set of \textit{edges}. Each edge connects a node $u\in U$ to another node $v\in V$ denoted as $(u,v)$. That is, we have $U\cap V=\emptyset$ and $E\subseteq U\times V$. Fig~\ref{bipartitenetwork} illustrates an example of a bipartite graph, where all nodes $u$ in $U$ are colored lime, and all nodes $v$ in $V$ are colored blue. It is evident that edges only exist between pairs 
% of nodes from different sets of nodes. 

% \textbf{Subgraph}: A bipartite graph $C_1 = (U_1, V_1, E_1)$ is a \textit{subgraph} of $C=(U,V,E)$ if both sets of nodes and the set of edges of $C_1$ are subsets of those of $C$, \textit{i.e.}, $U_1 \subseteq U, V_1 \subseteq V$, and $E_1 \subseteq E$.

% \textbf{bi-cliques}: A bipartite graph $C_1 = (U_1,V_1,E)$ is a \textit{bi-clique} of $C = (U,V,E)$ if $C_1$ is a \textit{subgraph} of $C$ and there is an edge between every node pair from different subsets of nodes in $C_1$, \textit{i.e.}, $U_1 \subseteq U, V_1 \subseteq V, E_1 \subseteq E$ and $U_1\times V_1=E_1$.

% \textbf{Maximal bi-cliques}: A bi-clique $C_1 = (U_1,V_1,E_1)$ is a \textit{maximal bi-clique} of a bipartite graph $C = (U,V,E)$ if it is not a sub-graph of any other bi-cliques of the bipartite graph, \textit{i.e.}, $\forall C_2 = (U_2,V_2,E_2)$ such that $U_1 \subseteq U_2 \subseteq U$ and $V_1 \subseteq V_2 \subseteq V$, $U_2\times V_2=E_2$ is satisfied if and only if $U_1 = U_2, V_1 = V_2$, and $E_1 = E_2$. Fig~\ref{bipartitenetwork} gives an example of maximal bi-clique. 

\subsection{Formal Concept Analysis (FCA)}\label{subsec2}

\textit{Formal concept analysis} (FCA) is a method for learning rules from a binary relational knowledge base, which is \textit{formal context}. Given a formal context, the aim of FCA is to extract \textit{formal concepts} and their hierarchical structure, which constitutes \textit{concept lattices}~(\cite{FCAbook,FCAintroduction}). 

\textbf{Formal Contexts}: A formal context is a triple $\mathbb K :=(G,M,I)$, where $G$ is a set of \textit{objects}, $M$ is a set of \textit{attributes}, and $I\subseteq G \times M$ is a binary relation called \textit{incidence}, expressing which object has which attribute. The relation is denoted by $gIm$ or $(g,m)\in I$, signifying that the object $g \in G$ possesses the attribute $m \in M$. 

\begin{figure}[!htbp]
    \centering
    \includegraphics[width=\textwidth]{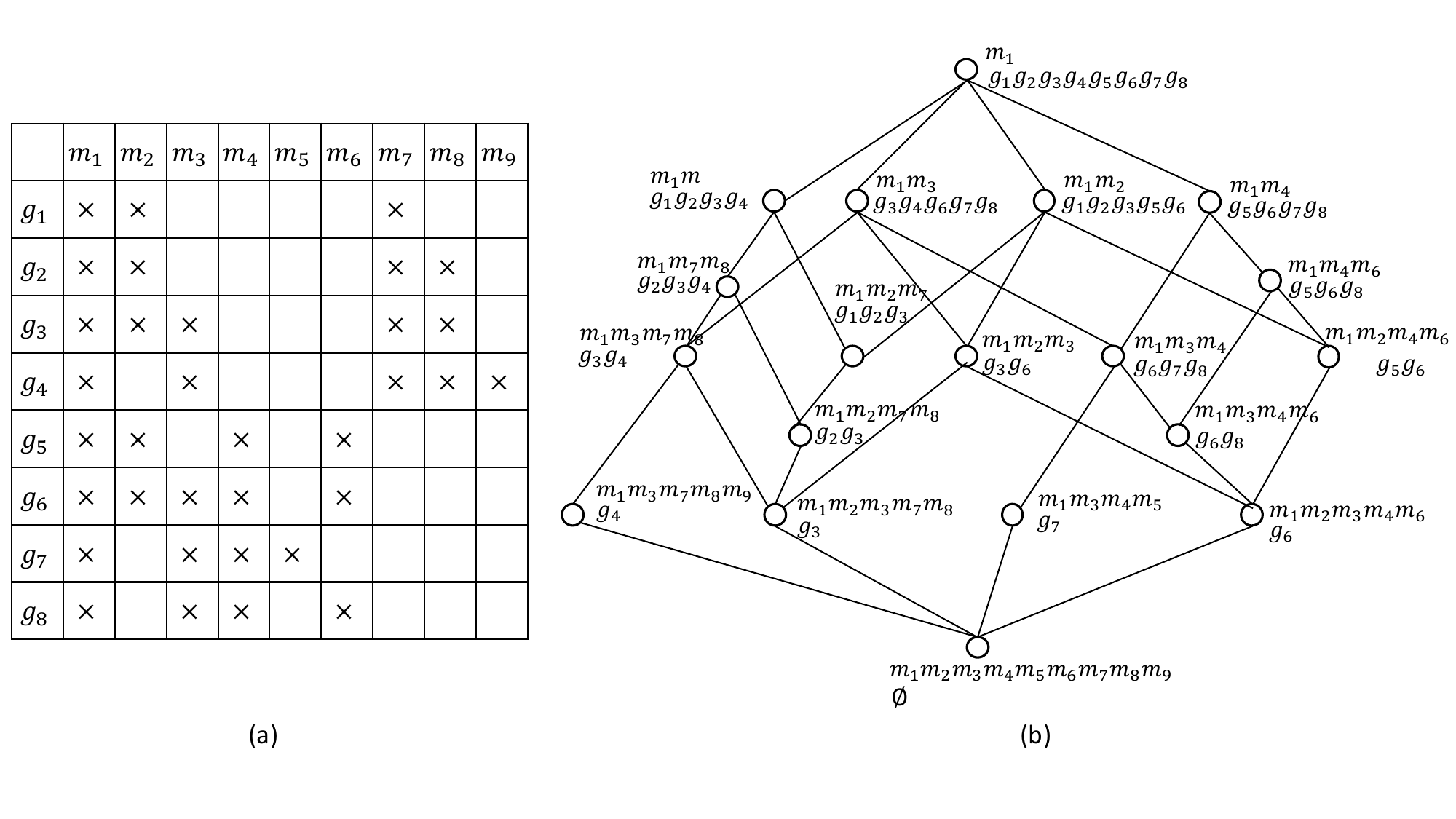}
    \caption{(a): A sample of formal context. (b): The concept lattice corresponds to the formal context in the left panel. \ }
    \label{conceptlattice}
\end{figure}

A formal context is illustrated in a binary table, as exemplified on the left of Figure~\ref{conceptlattice}(a), where each row represents an object and each column represents an attribute. A cell in the table is marked with a cross if the object in its row possesses the attribute in its column. 

\textbf{Formal Concepts}: In a context $\mathbb K =(G,M,I)$, for an object subset $A \subseteq G$ and an attribute subset $B \subseteq M$, $(A,B)$ is called a \textit{formal concept} if for each $(A_1, B_1)$ such that $A\subseteq A_1\subseteq G$ and $B\subseteq B_1\subseteq M$, $A_1\times B_1 \subseteq I$ is satisfied if and only if $A=A_1$ and $B=B_1$. If $(A, B)$ is a formal concept, $A$ is called an \textit{extent}, and $B$ is called an \textit{intent}. In a formal concept, the extent contains all objects that share all attributes in the intent, and the intent contains all attributes that are shared in common by all objects in the extent. 

\textbf{Concept Lattices}: Given a context $\mathbb{K} = (G, M, I)$, the \textit{concept lattice} of context $\mathbb{K}$, denoted by $\underline {\mathfrak B}(\mathbb{K})$, is the structure that organizes the set of all concepts extracted from context $\mathbb{K}$ with the \textit{hierarchical order} $<$. For two concepts $(A_1,B_1) $ and $(A_2,B_2)$, we write $(A_1,B_1) < (A_2,B_2)$ if $A_1\subset A_2$ (which mutually implies $B_2\subset B_1$)~(\cite{conceptlattice}). A concept lattice is visualized with a line diagram. For example, the line diagram depicted in~\ref{conceptlattice}(b) represents the concept lattice of the context represented in~\ref{conceptlattice}(a). In the diagram, nodes represent formal concepts and lines represent hierarchical orders.

\begin{figure}[!htbp]
    \centering
    \includegraphics[width=\textwidth]{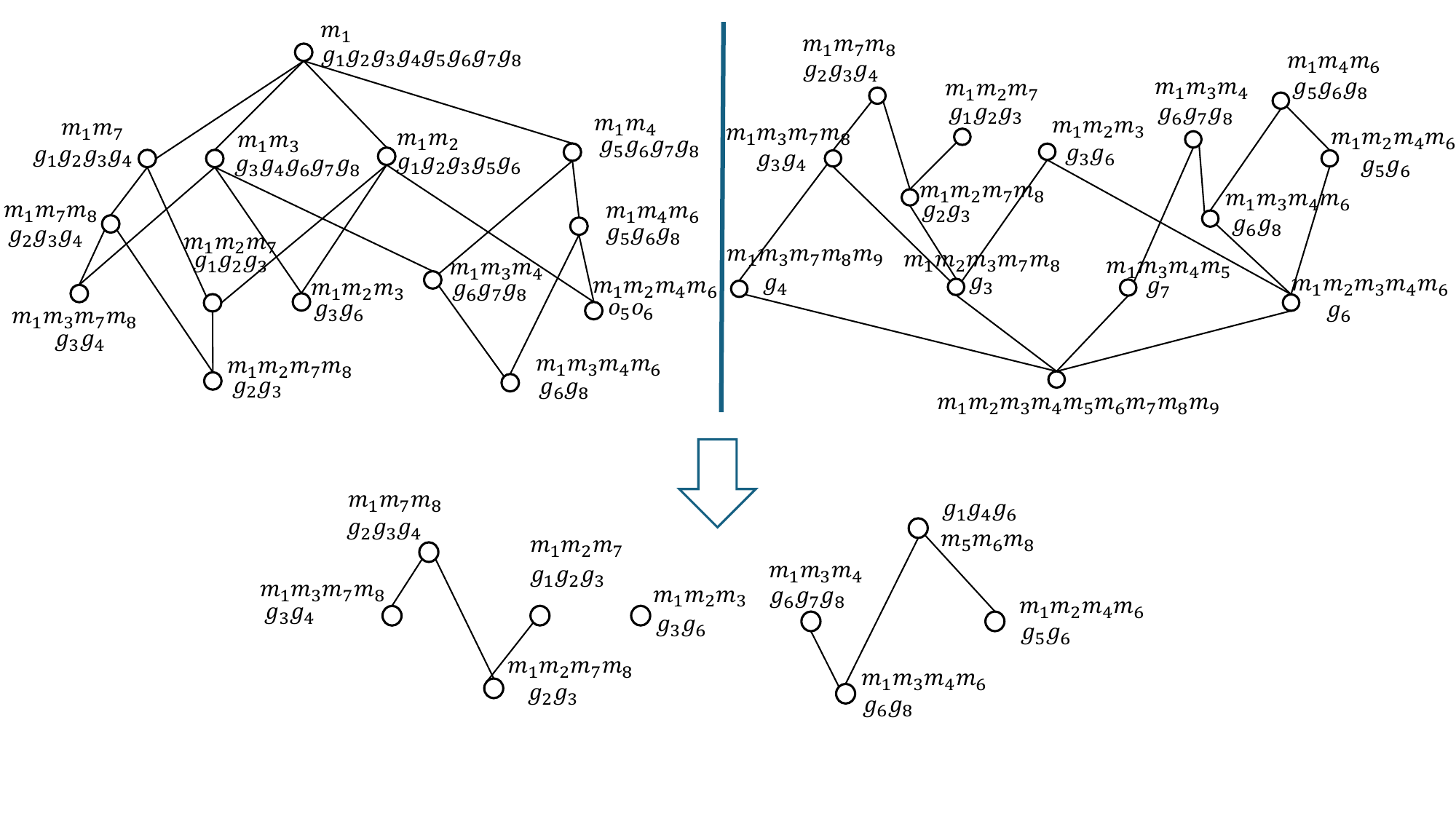}
    \caption{Top-left: The upward iceberg concept lattice of the above example is extracted by considering extents with a size of at least 2, or intents with a size of at most 4. Top-right: The downward iceberg concept lattice of the above example is extracted by considering extents with a size of at most 5, or intents with a size of at least 3. Bottom: The intersection of two iceberg concept lattices.\ }
    \label{icebergconceptlattice}
\end{figure}

\textbf{Iceberg Concept Lattice}: Given a context $\mathbb{K} = (G, M, I)$ and its concept lattice $\underline {\mathfrak B}(\mathbb{K})$, an \textit{iceberg concept lattice} $\underline {\mathfrak B}'(\mathbb{K})$ is defined to be the collection of all \textit{frequent concepts}, which are frequent closed sets, from the full concept lattice $\underline {\mathfrak B}(\mathbb{K})$. Here we say a concept is \textit{frequent} if its extent has a size larger than a pre-determined threshold $\theta$\footnote{Since objects and attributes are dual, in some research, a frequent concept may also be dually defined for the concepts whose intent is less than threshold $\theta$.}. Formally, an iceberg concept lattice is defined to be $\{(A_1,B_1)\in \underline {\mathfrak B}(\mathbb{K}) \vert \|A_1\|>\theta \}$. Figure~\ref{icebergconceptlattice} gives an example of iceberg concept lattices. 

As introduced above, in a concept lattice, the concepts are ordered based on the sizes of their extents. Hence, we can easily derive that an iceberg concept lattice is formed of concepts from the top-most part of the full concept lattice, which is the reason it is named an \textit{iceberg} concept lattice~(\cite{iceberglattice}). Since an iceberg concept lattice contains the concepts with the largest extents, it is often considered to represent the most ``significant'' part of the concept lattice, and is thus used in cases where the full concept lattice is considered redundant and needs to be reduced~(\cite{iceberglattice,frequent_itemset_mining}).

\subsection{FCA and Bipartite Networks}\label{subsec3}
\begin{figure}
    \centering
    \includegraphics[width=\textwidth]{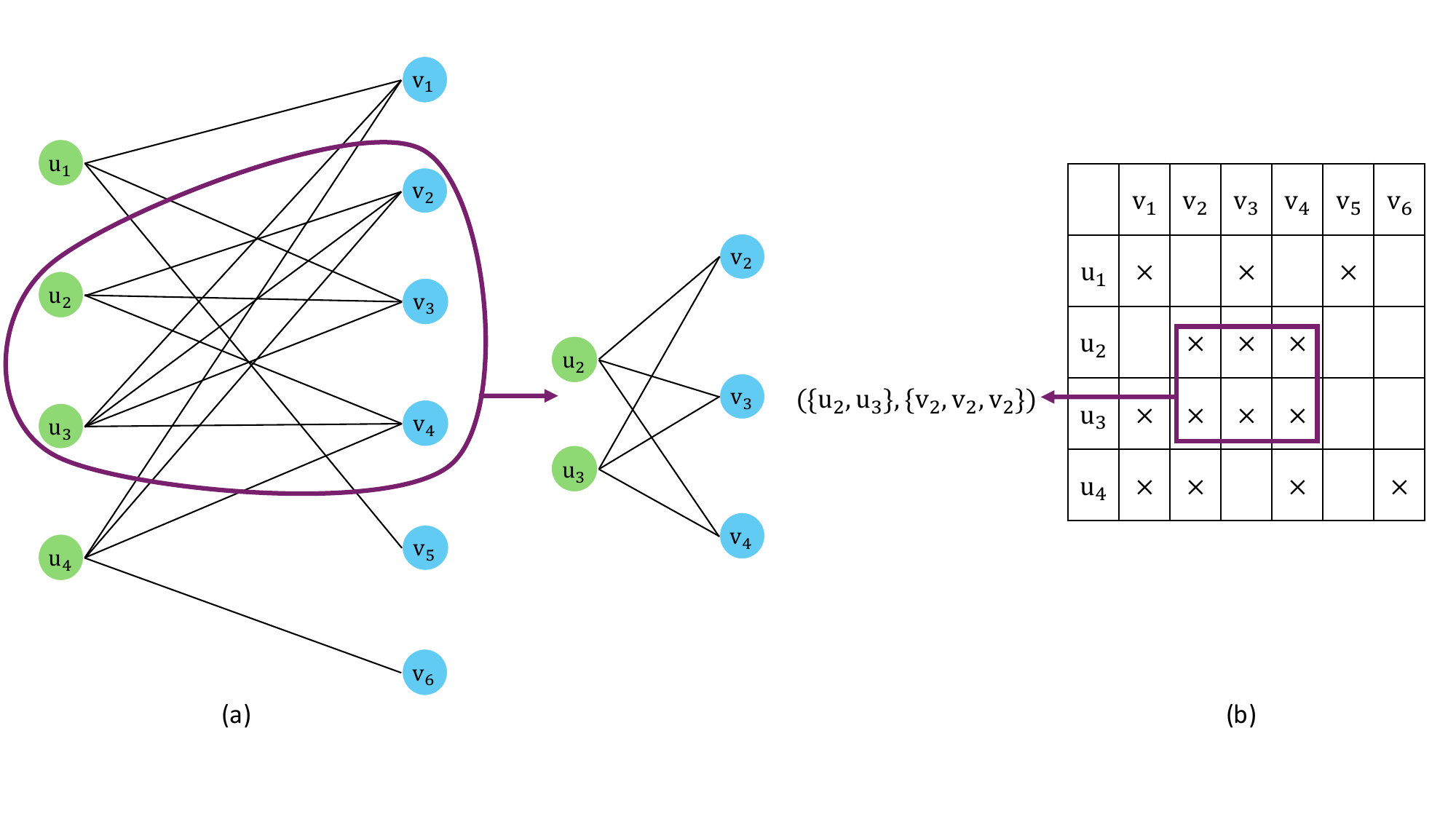}
    \caption{A depiction of the equivalence between bipartite networks and formal contexts, as well as the equivalence between maximal bi-cliques and formal concepts. The bipartite network in (a) can be represented as the formal context in (b). The sub-network circled in purple is a maximal bi-clique in the bipartite network in (a), which can be represented as a formal concept framed in the corresponding colors in the formal context in (b).}
    \label{FCAbi-clique}
\end{figure}

By comparing the definitions above, we can observe that the definition of a formal context is equivalent to that of a bipartite network, and the definition of a maximal bi-clique is equivalent to that of a formal concept. More precisely, for every bipartite network $C = (U, V, E)$, if we consider the two node sets of a bipartite network $U$ and $V$ as the object set and the attribute set, and the edge set $E$ as the binary relation of objects and attributes, we may easily find that $(U, V, E)$ should also be a formal context. Additionally, if $C_1=(A_1, B_1, E_1)$ is found to be a maximal bi-clique of $C$, it is certain that $(A_1, B_1)$ is also a formal concept in $\underline{\mathfrak B} (U, V, E)$. Fig~\ref{FCAbi-clique} provides an example of such an equivalence between bipartite networks and formal contexts, as well as the equivalence between maximal bi-cliques and formal concepts.

\subsection{Transformer Encoder}\label{subsec4}

\textit{Transformer}~(\cite{Transformer}) is a deep learning network structure known for its ability to capture dependencies between elements within a sequence. A full Transformer architecture consists of an \textit{encoder} and a \textit{decoder}. The encoder processes the input sequence and outputs an ``encoded'' sequence, while the decoder recursively generates the output sequence based on the input sequence, the ``encoded'' sequence generated by the encoder, as well as the elements previously output by the decoder itself.

Usually, the decoder should be used together with the encoder in order to form the traditional sequence-to-sequence structure, which takes a sequence as input and outputs another sequence. However, both the decoder and the encoder can be used separately for different applications. For example, the \textit{Generative Pre-training Transformer} (GPT)~(\cite{GPT}) is trained on a network with decoders only~(\cite{GPT}). On the other hand, the \textit{Bidirectional Encoder Representations from Transformers} (BERT)~(\cite{BERT}) model is trained on a network with encoders only~(\cite{BERT}). Generally, since the decoder works recursively, it is preferred in scenes where the local order of the elements in the input/output sequences is considered important. The encoder, on the other hand, is preferred in scenes where the global dependencies of all elements in the input/output sequences are considered important. In this research, we plan to use the encoder of the Transformer to learn the information of bi-cliques. 

% Since the elements in the bi-cliques are unordered, we consider the encoder-only Transformer more suitable for the task. Hence, here we will only give introductions to the structure of an encoder-only Transformer network. For more details on the structure of the Transformer decoder, please refer to~(\cite{Transformer}.

The detailed structure of the Transformer encoder architecture is illustrated in Fig~\ref{Transformerencoder}. It takes a sequence as its input, and the elements within the sequences are initially transformed into one-hot vectors before being fed into the network. These one-hot vectors are then further converted into dense vectors through the \textit{embedding layer}, enabling future processing. The embedding layer utilizes a full-connect layer to embed the tokenized one-hot vectors into the vector space where the dense vectors are located. Since the Transformer does not inherently understand the order of words in a sequence, a special mechanism called \textit{positional encoding} is incorporated through the embedding layers to capture information on the position of each token. This allows the network to distinguish between words based on their position in the sequence. After passing the embedding layer, the embeddings are fed into a stack of encoder layers, where each encoder layer is formed up with a \textit{multi-head attention} and a {feed-forward network. Finally, the output of the stacked encoders passes through a linear transformation layer to get the output of the whole transformer network.
\begin{figure}
    \centering
    \includegraphics[width=\textwidth]{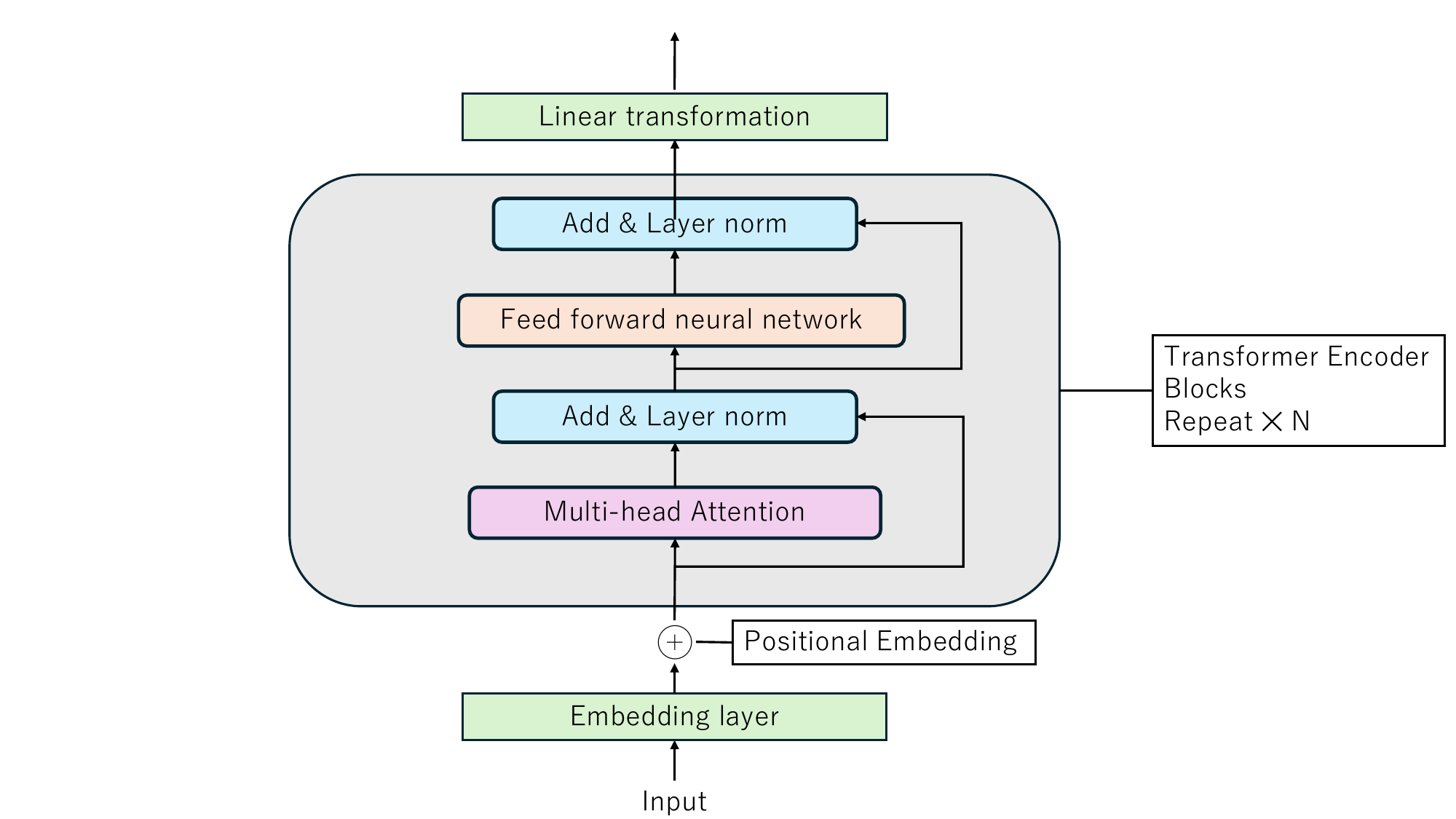}
    \caption{The detailed architecture of the standard Transformer encoder. Please refer to~(\cite{Transformer}) for the original figure.}
    \label{Transformerencoder}
\end{figure}

\subsection{The Transformer Encoder and FCA}\label{subsec5}

As mentioned above, the Transformer encoder takes sequential data as inputs like sentences and it processes the capability to capture the relationships between tokens within the sequence. Formal concepts, though lacking explicit order, can be treated as sequential data with an order-free syntax. By regarding objects and attributes as words in sentences, the extents and intents of a concept resemble sentences in a language with unordered syntax. We expect that the Transformer encoder can effectively capture dependencies between objects and attributes within the same formal concepts, which is equivalent to the dependencies of nodes within the maximal bi-cliques, as each formal concept corresponds to a maximal bi-clique. Since previous research has shown that such information can contribute to making link prediction work well, we decide to adopt the Transformer encoder for this task.

Given that the extents and intents of formal concepts are unordered, there is no need to keep track of the order, and in our method, the positional encoding mechanism is removed. Details on our modifications to the Transformer encoder will be introduced in Section~\ref{sec4}.

\section{Related Work}\label{sec3}

Based on the structural hole theory~(\cite{structurehole2}) which has been mentioned above, several rule-based methods, such as \textit{Missbin} and \textit{MF-NSS}~(\cite{missbin,matrixnega,linkbipartite}), have been proposed. These methods establish rules whereby if the overlap of two maximal bi-cliques exceeds a predefined threshold, the two bi-cliques merge into one, and the newly generated links are considered candidate links. These methods demonstrate better performance than some heuristic similarity-based methods such as Common Neighbors~(\cite{commonneighbors}) and Jaccard~(\cite{jaccard}).

However, manually crafted rules may not fully capture the information contained by bi-cliques. To better exploit the information of bi-cliques, researchers in~(\cite{FCA2VEC}) proposed \textit{FCA2VEC}, a method to embed the nodes into a vector space based on their co-occurrence relationship within the maximal bi-cliques. The co-occurrences of nodes in a maximal bi-clique can be converted to the co-occurrences of objects in a formal concept, resembling the co-occurrences of words in a sentence. Thus, one can use the embedding models similar to the well-known word embedding model \textit{Word2Vec}~(\cite{word2vec1}) to embed the nodes into vectors~(\cite{FCA2VEC}). The FCA2VEC method presents a novel strategy for node embedding, by converting networks into formal contexts, extracting information from maximal bi-cliques using FCA, and subsequently processing the extracted information using embedding methods. Besides such a Word2Vec-like embedding method, in~(\cite{boa}), the researchers proposed another method for node embedding based on bi-clique information using \textit{Bidirectional Long Short-Term Memory} (Bi-LSTM)~(\cite{bilstm}). Both methods have shown good performances and proven that the information extracted from bi-cliques is useful for link prediction~(\cite{boa,FCA2VEC}). However, these embedding methods cannot fully leverage all the information extractable by FCA, potentially limiting the prediction performance.  

To address this limitation, researchers in~(\cite{bert4fca}) introduced \textit{BERT4FCA}, a method utilizing \textit{BERT}~(\cite{BERT}) to comprehensively learn information extracted by FCA, encompassing all formal concepts and their order relations, and leverage it for link prediction. BERT4FCA outperforms previous bi-clique-based approaches~\cite{bert4fca}), underscoring the efficacy of incorporating richer information extracted by FCA for enhanced prediction performance. However, despite its remarkable prediction performance, it requires substantial memory and execution time. This problem renders it unsuitable for scenarios with limited computational resources. 

\section{Methodology}\label{sec4}

% To solve the previous bi-cliques-based bipartite link prediction methods' problem that they cannot be applied to large datasets, we introduce a novel method called \textit{BicliqueEncoder}, which is designed to handle large datasets with flexible adjustable computational resources. Its core ideology can be summarized in the following two points.
In this section we introduce a novel method called BicliqueEncoder, with which we can make link prediction in large datasets based on bi-cliques with flexible and adjustable computational resources. Our design policy can be summarized in the following two points.
\begin{itemize}
    \item Among all bi-cliques that can be extracted from the bipartite network, define the ``significant '' ones and develop a technique that can extract the significant bi-cliques only.  
    \item Use a technique that can effectively capture the information in these significant bi-cliques as well as the original bipartite network, in order to make precise link predictions.
\end{itemize}

For the first point, our method defines the significant bi-cliques to be those corresponding to concepts in the \textit{core iceberg concept lattice}. These significant bi-cliques can be extracted without exhaustive \textbf{complete} enumeration of all formal concepts and are expected to be indeed significant for link prediction. For the second point, we have developed a modified Transformer encoder network called \textit{unordered Transformer encoder} to effectively capture the information and make link predictions. We will first give a brief introduction to these techniques we develop before giving the detailed working flow of our method. 

In the rest part of the paper, to keep it concise, when we use terms in FCA, we will no longer explain their equivalences in bipartite network theory. For example, we will directly say ``extracting significant formal concepts'' instead of ``extracting formal concepts corresponding to significant bi-cliques''. Readers who are not familiar with such equivalences can refer back to the preliminaries in the last section.

\subsection{Core iceberg concept lattice}\label{subsec2}

Currently, there have been many different definitions for significant formal concepts and methods for extracting them. However, for most definitions, the extraction methods only work in a selective mode, that is, they need to exhaustively enumerate all formal concepts and then ``pick out'' the significant ones. The time complexity of such an exhaustive enumeration process is lower bounded by $O(C)$, where $C$ represents the number of all formal concepts. In the worst scenario, the size of a concept lattice will be exponential to the number of objects, attributes and incidences, making such a process extremely time-consuming. Therefore, the only effective way to reduce the time caused by extracting formal concepts is to choose a definition for significant formal concepts so that we can directly extract the significant concepts without the exhaustive enumeration of all concepts.

In our method, we define a concept $(A,B)$ to be significant if $l_1\leq \|A\| \leq u_1$ and $l_2\leq \|B\| \leq u_2$. That is, a concept is significant if both its intents' and extents' sizes are neither too long nor too short. According to the definition of an iceberg concept lattice, we can find that:
\begin{itemize}
    \item All concepts $(A, B)$ such that $|A|\geq l_1 $ and $|B|\leq u_2 $ form an ``upward'' iceberg concept lattice.
    \item All concepts $(A, B)$ such that $|A|\leq u_1$ and $|B|\geq l_2$ form a ``downward'' iceberg concept lattice.
    \item The significant concepts are those in the intersection of the upward iceberg and the downward iceberg. We name this intersection the ``core iceberg concept lattice''.
\end{itemize}
Since the concepts in the iceberg concept lattice can be extracted without exhaustive enumeration by algorithms like LCM~(\cite{LCM}), which is a frequent closed sets mining algorithm. By applying a simple modification to the LCM algorithm, we will be able to extract the significant concepts in the core iceberg concept lattice without exhaustive enumeration. Details will be introduced in the workflow of our method.

Certainly, if the significant concepts are only easy to extract, but are not really significant in terms of the information they can provide for link prediction, extracting them would be meaningless. However, we consider the concepts in the core iceberg concept lattice to be indeed significant for link prediction because they encompass objects and attributes that frequently appear in concepts, as well as their binary relations~(\cite{iceberglattice}).

Nevertheless, no matter how significant are those concepts, they still only represent a subset of all concepts, it's possible that some objects and attributes may not be included within these concepts. To ensure the information of all objects and attributes is captured by the method, in addition to those extracted significant concepts, we also create training samples directly generated from the raw context. Details with be introduced in the workflow of our method.

\subsection{Unordered Transformer Encoder}\label{subsec3}

Previous research has utilized many different methods for capturing the information from formal concepts, including pure rule-based method~(\cite{missbin}), matrix-factorization-based methods~(\cite{matrixnega}), and deep-learning-based methods~(\cite{FCA2VEC,boa,bert4fca}). While in this research, we consider the Transformer encoder most suitable for the following reason. The most valuable information a formal concept can provide is the closed relations between objects and attributes within its extent and intent~(\cite{FCAbook}). The Transformer network is designed for extracting dependencies between elements in a sequence, so we believe it can capture the information provided by a formal concept well.

As mentioned in the previous section, we consider an encoder-only network more suitable for this task, because the decoder works recursively and thus it strongly relies on the order of the elements in the sequence, while formal concepts are completely unordered. Furthermore, the original version of the Transformer encoder still has a mechanism called \textit{positional embedding}, which memorizes the order of elements in the input sequence. Since the extents and intents are completely unordered, in our method, we have removed this positional encoding mechanism. We name our modified Transformer architecture an \textit{unordered Transformer encoder} and believe it is well suitable for our task.

\subsection{The workflow of our method}\label{subsec4}

Our method consists of two steps: \textit{data preparation} and \textit{training}. An overview of the workflow of our method is shown in Fig.~\ref{ConceptEncoder}. 
\begin{figure}
    \centering
    \includegraphics[width=\textwidth]{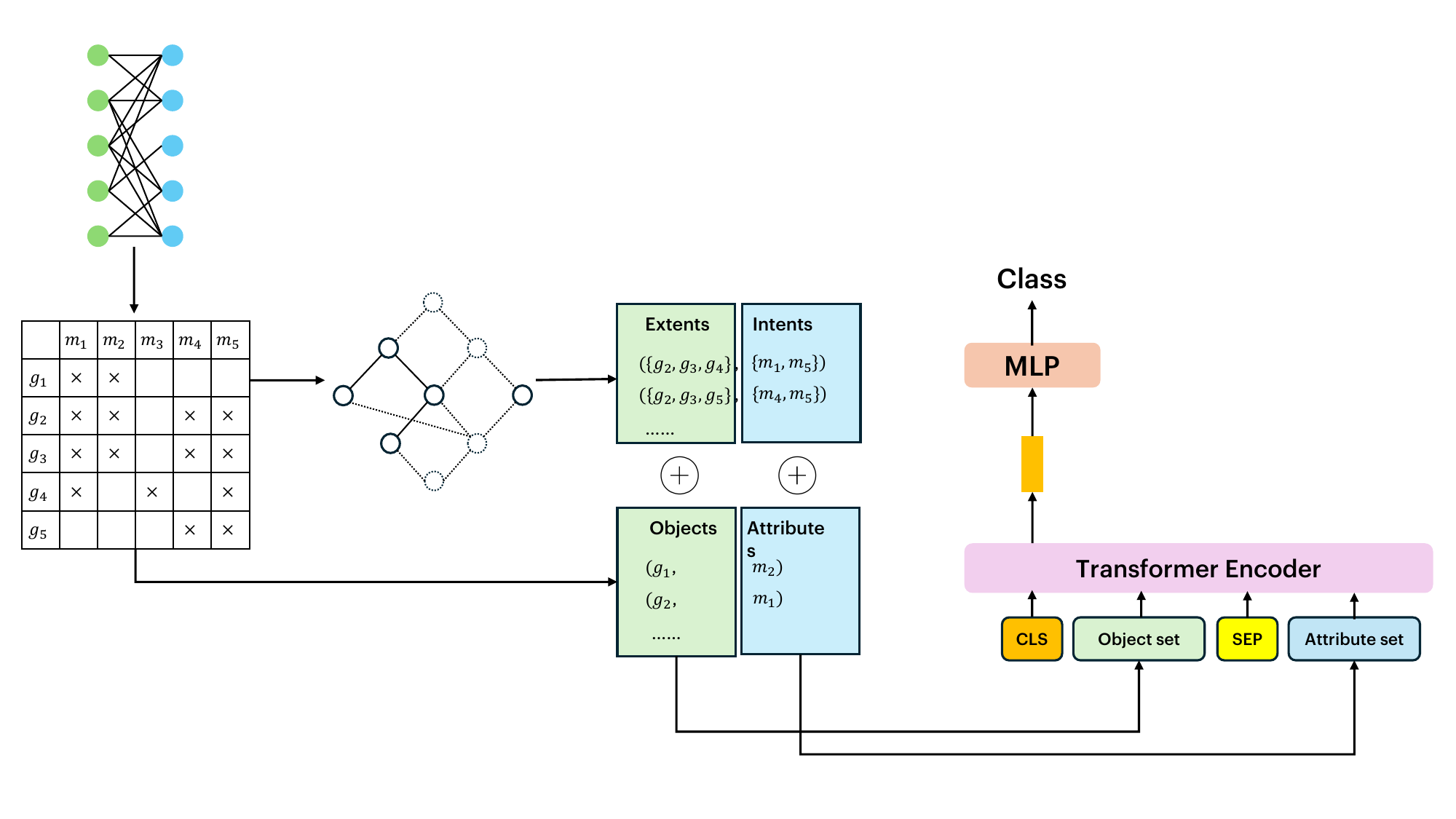}
    \caption{An overview of the working flow of our method.}
    \label{ConceptEncoder}
\end{figure}

\textbf{Significant Concept Extraction}: In this step, we convert the bipartite network into a formal context. Then, we set the upper and lower bounds on the sizes of extents and intents, \textit{i.e.}, $u_1,u_2,l_1$, and $l_2$, and only extract concepts in which the sizes of extents and intents fall in this range. That is, we are to search all significant concepts in the core iceberg concept lattice. Such a searching process is achieved by an algorithm derived from the \textit{LCM} algorithm~(\cite{LCM}). Please refer to~\autoref{LCM}} for a detailed algorithm.

The upper and lower bounds for the sizes of extents and intents should be determined based on the size and density of formal contexts, as well as the available computational resources. If the bounds are too restrictive, there may be no concepts in the core iceberg concept lattice. On the other hand, too loose bounds may result in an excessive number of extracted concepts, which can be computationally expensive. In practice, the choice of bounds involves a trade-off between prediction performance and computational resources. 

\textbf{Training Data Preparation}: With the significant concepts extracted, we are to generate the training samples. The training samples we use in our method consist of the \textit{concept samples} and the \textit{context samples}. The concept samples are generated from the significant concepts, and the context samples are generated from the original input context. Here we only briefly introduce data preparing procedure. Please refer to~\autoref{data preparation} for the details of this procedure. 

Before generating the concept samples, we need to generate four intermediate sets. The extents are collected into a set $E_\mathrm{p}$ and all intents are collected into a set $I_\mathrm{p}$. Then, we are to generate two distractor sets $E_\mathrm{n}$ and $I_\mathrm{n}$, representing non-extent object sets and non-intent attribute sets, correspondingly. The details and the corresponding pseudocode are presented in~\ref{Appdedix B1}.

After generating the four intermediate sets, we can generate the concept samples. The positive samples consist of all formal concepts, which are pairs of an extent and an intent, denoted as $C_\mathrm{p}$. The negative samples are generated from non-extent object sets $E_\mathrm{n}$ and non-intent attribute sets $I_\mathrm{n}$, denoted as $C_\mathrm{n}$. The details and the corresponding pseudocode are presented in~\ref{Appdedix B2}.

Besides the concept samples, we still need to generate the context samples because the concept samples may leave out some information of some minor attributes or objects. The context samples are generated from the original formal context. For every $g \in G$ and $m \in M$ such that $(g, m) \in I$, we add the pair $(\{g\}, \{m\})$ as a positive sample and add it to $T_p$. Then we randomly select a pair of an object $g_1 \in G$ and an attribute $m_1 \in M$, if $(g_1,m_1) \notin I$, we treat this pair as a negative sample and add it into $T_n$. The details and the corresponding pseudocode are presented in~\ref{Appdedix B3}. 

After obtaining all training samples, we need to pad them to the same length with special tokens ``[PAD]'' so that they can be fed into the network. We find the length of the longest sequence and pad all sequences to this length. The details and the corresponding pseudocode are presented in~\ref{Appdedix B4}. After this step, we will obtain the final training set $T_\mathrm{F}$, with its label (positive or negative) set.

\begin{figure}[H]
    \centering
    \includegraphics[width=\textwidth]{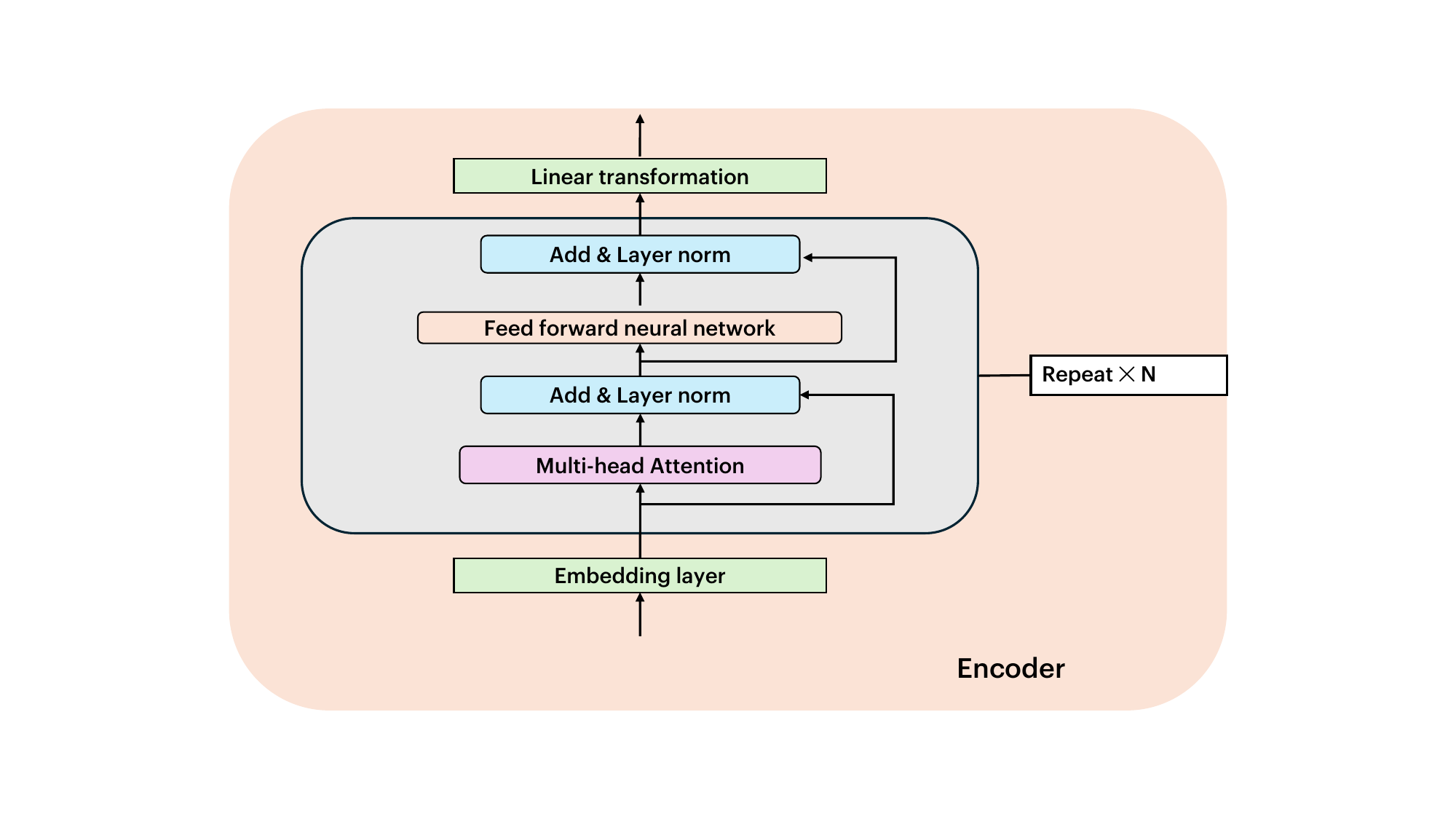}
    \caption{The detailed architecture of the Transformer encoder used in our method. We remove the position embedding from the standard Transformer encoder which is illustrated in Fig~\ref{Transformerencoder}. }
    \label{Encoder}
\end{figure}

\textbf{Training}: We employ an unordered Transformer encoder and a Multi-Layer Perceptron (MLP) to train a model for link prediction. The model takes concatenations of object sets and attribute sets as input. A special token [CLS] is added at the first position of an input sequence. After the training step, the final embedding of this [CLS] token is expected to capture and represent the information of the whole sequence. Another special token [SEP] is added between object sets and attribute sets to help the model differentiate two different sets. 
The detailed structure of the Transformer encoder model is illustrated in Fig~\ref{Encoder}. 

In our experiments, the model contains $N=9$ identical encoder layers, each of which contains a multi-head attention layer with $H=12$ heads and a feedforward neural network. The input size is $(n_{\mathrm{obj}} + n_{\mathrm{attr}} )\times (l_{\mathrm{ext}} +  l_{\mathrm{int}} )$, where $n_{\mathrm{obj}}$ is the number of objects, $n_{\mathrm{attr}}$ is the number of attributes, $l_{\mathrm{ext}}$ is the size of the longest extents, and $l_{\mathrm{int}}$ is the size of the longest intent. 

\begin{figure}
    \centering
    \includegraphics[width=\textwidth]{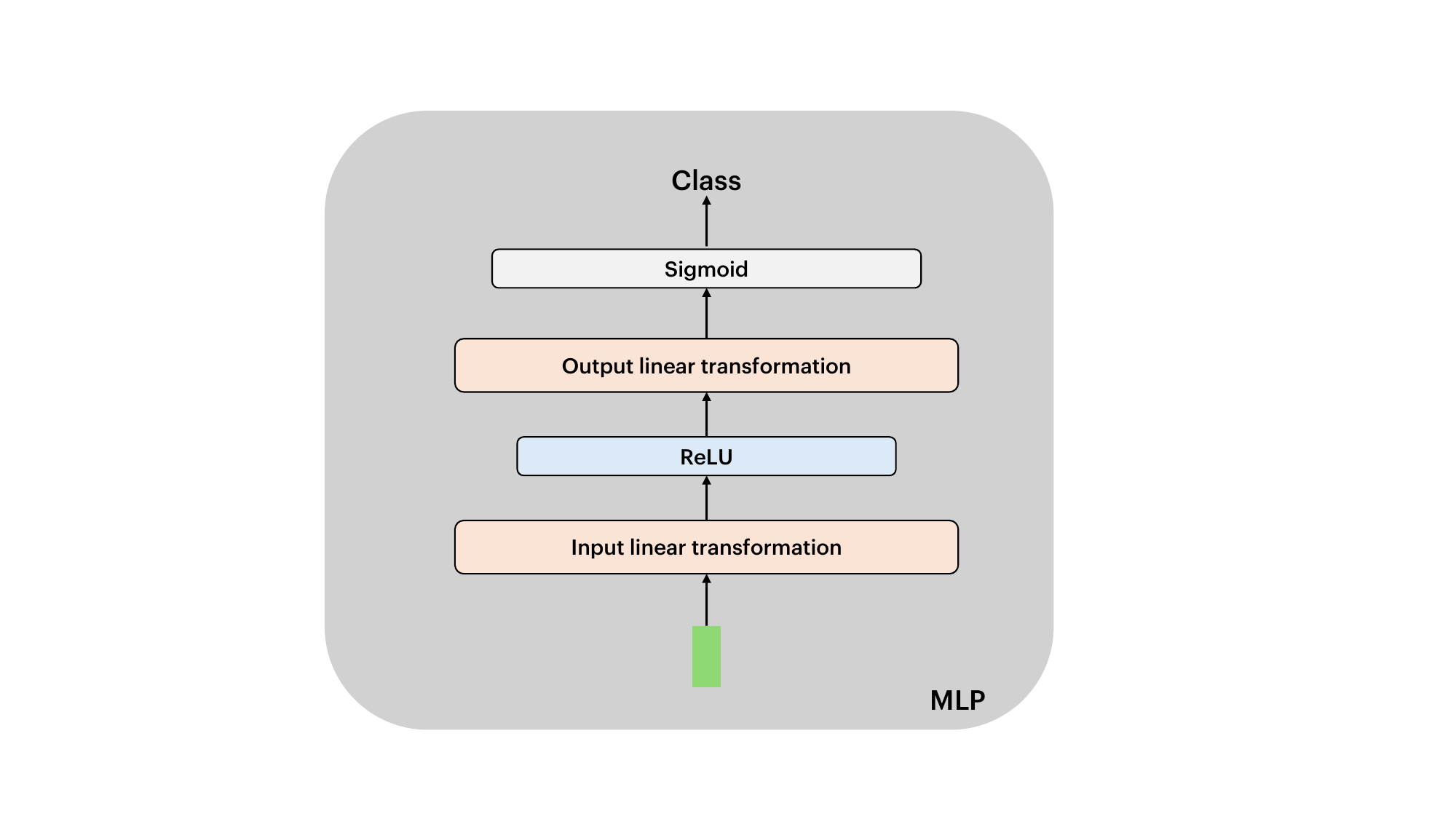}
    \caption{The detailed architecture of the MLP.}
    \label{MLP}
\end{figure}

We take the final hidden state of the special first token [CLS] as the output of the encoder for each input sequence, and the size of the output vector is set to $d_{\mathrm{model}} = 768$. Subsequently, this vector is passed through an MLP consisting of a hidden layer with the size of $512$ and a sigmoid function to output 0 or 1. The detailed structure of MLP used in our model is illustrated in Fig~\ref{MLP}.

\newpage
With the network for training setup, we are to feed the training samples into the network to train the model. For each sample $(X,Y)$ from the final training set $T_\mathrm{F}$, ``$[CLS] + X + [SEP] +  Y $ '' is fed into the model. The training process may be conducted multiple times (epochs) until the loss has converged.

\subsection{Discussion of our method}\label{subsec6}

Above is the whole workflow of our method. We can see that our method has two advantages over the previous bipartite link prediction methods utilizing FCA. 

First, in contrast to previous FCA-based methods, our method is applicable to large datasets with the help of the core iceberg concept lattice. While previous FCA-based methods require the full concept lattice for link prediction, our method selects significant concepts to avoid the requirement for an exhaustive enumeration of all concepts. For example, if the number of formal concepts in a formal context is $N$, the time complexity of the FCA concept extraction process in other FCA-based methods is $O(N)$, whereas in our method, it is $O(n)$, where $n$ is significantly smaller than $N$. Furthermore, in extreme cases where the extracted significant concepts provide limited useful information for link prediction, our method can still learn directly from the raw formal context and achieve strong performance, enhancing its flexibility and robustness.

Second, our method offers flexibility in adjusting computational resources according to specific needs. As mentioned above, there is a trade-off between prediction performance and computational resources. Extracting a greater number of concepts typically enhances prediction performance but also increases the time required for concept extraction and model training. Additionally, longer extents and intents will lead to a larger model size, which will require more computational resources during training. One can adjust the number and the length of extracted concepts by defining the range of sizes for extents and intents based on the available computational resources. This adaptability ensures efficient resource utilization while maintaining competitive prediction performance.  

Additionally, in our method, we propose a Transformer network which is specifically for bipartite link prediction. We remove the ``positional embedding'' and add special tokens ``[CLS]'' and ``[SEP]'' to make the Transformer architecture suitable for tasks in bipartite networks so that it can directly and effectively learn the pairwise relationships between two nodes from different sets in the bipartite networks. With the help of the powerful attention mechanism in the Transformer encoder networks, the negative influence on prediction performance caused by the information not contained in the extracted formal concepts can be counterbalanced. Moreover, the proposed network can directly process bipartite networks to make link predictions with high performance. We will demonstrate the performance of it in the ablation experiments. 

However, our method also has some limitations. Focusing only on significant bi-cliques may result in the loss of minor patterns that are not contained in the extracted bi-cliques. Bi-cliques represent clusters of nodes that are connected to each other. They are important for the model to uncover the patterns of link formation underlying bipartite networks, as they help the model learn to identify pairs of nodes that connect (positive samples). However, nodes that do not connect are also important for link prediction, as they help the model learn to identify pairs of nodes that do not connect (negative samples). Only learning from bi-cliques may not effectively help the model identify negative samples. 

The importance of bi-cliques for link prediction also varies across different datasets, leading to varying improvements in prediction accuracy due to the use of bi-cliques. In some datasets, bi-cliques may perfectly reflect the pattern of link formation, resulting in high prediction accuracy, while in others, they may not capture the underlying structure as effectively.

\section{Experiments and Results}\label{sec5}

\subsection{Datasets}\label{subsec1}
We conduct experiments on five real-world datasets: \textit{CTD}, \textit{BMS-POS}, \textit{HetRec}, \textit{MovieLens}, and \textit{Bonanza}. These datasets are from various domains to verify the practicality of our method in real-world scenarios. We depict the features of these datasets in ~\autoref{description}. A detailed description of each follows. 

\begin{table}[!htbp]
% \begin{adjustwidth}{-2.25in}{0in}
    \centering
    \caption{\bf The features of the three datasets.}
    \begin{tabular}{ccccccc}
    \hline
         Dataset        &  Objects&  Attributes&  Incidences&  Extracted Concepts\\
    \hline
         CTD            & 10225 & 3283 & 103845 & 16702 \\

         BMS-POS        & 10000 & 1004 & 70522 & 5247  \\

         HetRec        & 1872 & 3261 & 143010 & 11508 \\

         MovieLens     & 8188 & 5995 & 469373 & 13458 \\

         Bonanza       & 7919 & 1973 & 36543 & 13380  \\
                                      
    \hline
    \end{tabular}
    \label{description}
% \end{adjustwidth}
\end{table}

\textbf{CTD}: This dataset represents a chemical-disease interaction network and is available at https://ctdbase.org. In this network, the objects correspond to chemicals, the attributes represent diseases, and each incidence indicates that a particular chemical is effective against a specific disease. The objective is to simulate a real-world scenario where we aim to predict future interactions based on the existing network. To accomplish this, we utilize the original network as the target network and generate the input network by randomly removing $10\% $ of the chemical-disease edges.
 
\textbf{BMS-POS}: This dataset represents product purchased transaction records provided by KDD CUP 2000, which is available at https://kdd.org/kdd-cup/view/kdd-cup-2000. In this bipartite network, the objects represent products, the attributes correspond to purchasing transactions, and each incidence indicates that a particular product was bought in a specific purchasing transaction. Our objective is to simulate a practical scenario where certain parts of the network are missing, and we aim to leverage the available network data to predict the missing edges. Similar to the approach used for the CTD dataset, we utilize the original network as the target network and generate the input network by randomly removing $10\%$ of the product-transaction edges.

While the BMS-POS dataset has been previously utilized in BERT4FCA, we employ the entire BMS-POS dataset for our experiments. In contrast, BERT4FCA use only a part of it, as the entire dataset is too large for BERT4FCA to process.

\textbf{HetRec}: The HetRec dataset~(\cite{HetRec}) represents a user-movie rating dataset collected from different websites. Originally, the dataset consisted of users rating movies on a scale of 1 to 5. To transform it into a binary dataset, we consider the scores equal to or greater than 3 as indications of user preferences(favorites), while the scores below 3 represent dislikes. This conversion transforms the original dataset into a binary user-movie dataset. In this bipartite network, users represent one set of nodes, movies represent the other set, and an edge denotes a user liking a particular movie. Our objective is to simulate a scenario where we predict movies that users may like in the future, akin to a movie recommendation system. We follow the same methodology used for CDT and BMS-POS datasets, utilizing the original network as the target for prediction and generating the input network by randomly removing $10\%$ of user-movie edges.

\textbf{MovieLens}: This dataset, like HetRec, is also a user-movie rating dataset generated from the MovieLens 25M dataset~(\cite{movielens}). We consider the scores equal to 5 as indications of user preferences, while the scores below 5 represent dislikes to transform the original dataset into a binary relational dataset. Then we remove the users who rated fewer than 300 movies and the movies which are rated by fewer than 120 users from the datasets. The objective and the methodology to generate the input and target networks are the same as those used in the HetRec dataset. 

\textbf{Bonanza}: This dataset represents buyer-seller rating records collected from the e-commerce website Bonanza. Buyers purchase products and rate the sellers with ``Positive'', ``Neutral'', or ``Negative'' scores. To generate a binary relational dataset from the original dataset, we treat buyers as one set, sellers as another set of nodes, and the rating records of ``Neutral'' or ``Negative'' as present links. Link prediction on such a dataset can help sellers identify possible negative reviews. We follow the same methodology used in the above datasets, utilizing the original network as the target for prediction and generating the input network by randomly removing $10\%$ of buyer-seller edges.

Above is a detailed description of all five datasets. We do not know the number of concepts in each dataset because determining the number of concepts requires extracting all concepts, which takes several days. Previous FCA-based methods face challenges when applied to these datasets because they require the entire concept lattice. 

The generation of test samples is as follows. We treat the removed incidences as positive samples. Then we generate the negative samples using the same method as the generation of training context samples. The set of test negative samples is denoted as $S_n$. We randomly select a pair of an object that $g_1 \in G $ and an attribute that $m_1 \in M$. If $(g_1,m_1) \notin I$, $(g_1,m_1) \notin S_n$, and $(g_1,m_1) \notin T_n$(the set of negative context sample pairs), then we treat it as a negative sample pair and add it in $S_n$. To ensure a balanced test set, we maintain the same number of negative samples as positive samples. 
% The same as the training step, during testing, all objects in positive and negative pairs are input to the extents encoder and all attributes in positive and negative pairs are input to the intents encoder. Each object and attribute are input simultaneously.

\subsection{Experiments}\label{subsec2}

The baseline methods we compare our method with are as follows:  several heuristic rule-based link prediction methods like \textit{Common Neighbors} (CN)~(\cite{commonneighbors}), \textit{Adamic-Adar Coefficient} (AA)~(\cite{linkbipartite}), \textit{Jaccard Coefficient} (JC)~(\cite{JC}), \textit{Resource Allocation}(RA)~(\cite{RA}), \textit{Rooted PageRank}(RPR) and \textit{SimRank}(SR)~(\cite{SR}); a classic bipartite link prediction method, \textit{Matrix factorization with singular value decomposition}(MF-SVD); an embedding-based method, \textit{random walk}, which trains node embedding using a random walk process and then trains a linear regression classifier on samples of present links with the features of nodes set to their embeddings to make predictions; a novel GNN-based link prediction method, \textit{SBGNN}, which is specifically designed for bipartite network~(\cite{SBGNN}). The results are reported in~\autoref{main}.

\begin{sidewaystable}[!htbp]
% \begin{adjustwidth}{-2.25in}{0in}
    \small
    \centering
    \caption{\bf The results for link prediction in bipartite networks.}
    \begin{tabular}{c|p{0.8cm}p{0.8cm}p{0.8cm}|p{0.8cm}p{0.8cm}p{0.8cm}|p{0.8cm}p{0.8cm}p{0.8cm}|p{0.8cm}p{0.8cm}p{0.8cm}|p{0.8cm}p{0.8cm}p{0.8cm}}
    % \begin{tabular}{c|ccc|ccc|ccc|ccc|ccc}
    \toprule
    \multirow{2}{*}{Method} & \multicolumn{3}{c|}{CTD} & \multicolumn{3}{c|}{BMS-POS} & \multicolumn{3}{c|}{HetRec} & \multicolumn{3}{c|}{MovieLens} & \multicolumn{3}{c}{Bonanza} \\
    & $F_1$ & AUC & AUPR &  $F_1$ & AUC & AUPR & $F_1$ & AUC & AUPR & $F_1$ & AUC & AUPR & $F_1$ & AUC & AUPR \\
    \midrule
     CN               & 0.696 & 0.756 & 0.748 & 0.670 & 0.715 & 0.712 & 0.697 & 0.745 & 0.735 & 0.688 & 0.735 & 0.727 & 0.187 & 0.522 & 0.516     \\
     AA               & 0.696 & 0.758 & 0.761 & 0.588 & 0.721 & 0.734 & 0.697 & 0.749 & 0.744 & 0.687 & 0.741 & 0.739 & 0.035 & 0.523 & 0.527     \\
     JC               & 0.438 & 0.727 & 0.649 & 0.611 & 0.562 & 0.471 & 0.682 & 0.641 & 0.609 & 0.622 & 0.560 & 0.531 & 0.064 & 0.519 & 0.492     \\
     RA               & 0.525 & 0.758 & 0.761 & 0.593 & 0.719 & 0.721 & 0.716 & 0.755 & 0.748 & 0.700 & 0.750 & 0.743 & 0.007 & 0.523 & 0.529     \\
     RPR         & 0.767 & 0.840 & 0.800 & 0.760 & 0.878 & \textbf{0.878} & 0.680 & 0.726 & 0.719 & 0.614 & 0.731 & 0.729 & 0.512 & 0.773 & 0.757     \\
     SR          & 0.662 & 0.546 & 0.514 & 0.001 & 0.251 & 0.357 & 0.529 & 0.431 & 0.420 & 0.501 & 0.416 & 0.451 & 0.010 & 0.418 & 0.429     \\
     MF-SVD           & 0.872 & \textbf{0.941} & 0.687 & 0.851 & \textbf{0.919} & 0.659 & 0.798 & \textbf{0.867} & 0.602 & 0.837 & \textbf{0.909} & 0.641 & 0.768 & 0.827 & 0.580 \\ 
     Random walk      & 0.622 & 0.572 & 0.552 & 0.522 & 0.529 & 0.503 & 0.706 & 0.707 & 0.646 & 0.487 & 0.533 & 0.528 & 0.622 & 0.506 & 0.490     \\
     SBGNN-GAT        & 0.867 & 0.867 & 0.819 & 0.842 & 0.840 & 0.784 & 0.839 & 0.850 & 0.774 & 0.895 & 0.893 & 0.843 & 0.834 & 0.832 & 0.773     \\
     SBGNN-MEAN       & 0.866 & 0.867 & 0.820 & 0.849 & 0.847 & 0.790 & 0.842 & 0.838 & 0.776 & 0.898 & 0.896 & \textbf{0.847} & 0.836 & 0.833 & 0.773     \\
     BicliqueEncoder  & \textbf{0.916} & 0.933 & \textbf{0.883} & \textbf{0.915} & 0.909 & 0.850 & \textbf{0.895} & 0.866 & \textbf{0.819} & \textbf{0.906} & 0.899 & 0.838 & \textbf{0.892} & \textbf{0.882} & \textbf{0.814}     \\
    \bottomrule
    \end{tabular}
    \begin{flushleft}
        SBGNN-GAT and SBGNN-Mean are two versions of SBGNN which follow different message passing schemes.
    \end{flushleft}
    \label{main}
% \end{adjustwidth}
\end{sidewaystable}

From the results, we can observe that BicliqueEncoder has overall better performances than other baseline methods across all five datasets. For MF-SVD, while it exhibits higher AUC scores than BicliqueEncoder across all datasets, its overall performance is still considered lower than BicliqueEncoder because of its low AUPR scores. According to previous research, the AUPR score gives more weight to the positive samples. A low AUPR score indicates that the model's high prediction scores do not correlate well with being in the positive class, which suggests that the model has difficulty achieving high precision~(\cite{aucaupr}). Since in link prediction we focus on predicting the generation of new links, which is predicting the positive samples but not the negative samples~(\cite{bipartite1,matrixnega}), so if a model exhibits a low AUPR, it is considered to have a low performance in link prediction. 

From the results, we can also find that the heuristic methods' performances are unstable across different datasets. For example, rooted PageRank has the highest AUPR score on the BMS-POS dataset among all methods, while medium AUPR score on other datasets, and all heuristic methods except rooted PageRank perform badly on the Bonanza dataset. The unstable performance of heuristic methods is due to the pre-defined similarity score computing methods working well in some networks, while working badly in others. BicliqueEncoder, on the other hand, has stable good performance across different datasets. 

\subsection{Ablation Experiment}\label{subsec3}
To further analyze our method, we conduct an ablation experiment to evaluate whether adding concepts as training samples indeed contributes to link prediction performance. 

In the ablation experiment, we train the model using only context training samples and then evaluate the prediction performance. By comparing the results of the model trained with and without concepts, we can assess the contribution of the information learned from concepts, and evaluate the capability of our proposed model for extracting useful information for link prediction from raw bipartite networks. The results are reported in~\autoref{ablation}.

\begin{sidewaystable}[!htbp]
    \small
    \centering
    \caption{\bf The results for the first ablation experiment.}
    % \begin{tabular}{c|ccc|ccc|ccc|ccc|ccc}
    \begin{tabular}{c|p{0.8cm}p{0.8cm}p{0.8cm}|p{0.8cm}p{0.8cm}p{0.8cm}|p{0.8cm}p{0.8cm}p{0.8cm}|p{0.8cm}p{0.8cm}p{0.8cm}|p{0.8cm}p{0.8cm}p{0.8cm}}
    \toprule
         \multirow{2}{*}{Method} & \multicolumn{3}{c|}{CTD} & \multicolumn{3}{c|}{BMS-POS} & \multicolumn{3}{c|}{HetRec} & \multicolumn{3}{c|}{MovieLens} & \multicolumn{3}{c}{Bonanza} \\
    \cmidrule{2-16}
         & $F_1$ & AUC & AUPR & $F_1$ & AUC & AUPR & $F_1$ & AUC & AUPR & $F_1$ & AUC & AUPR & $F_1$ & AUC & AUPR \\
    \midrule
         BicliqueEncoder         & 0.916 & 0.933 & 0.883 & 0.915 & 0.909 & 0.850 & 0.895 & 0.866 & 0.819 & 0.906 & 0.899 & 0.838 & 0.892 & 0.882 & 0.814     \\
         BicliqueEncoder-NC      & 0.886 & 0.883 & 0.837 & 0.908 & 0.902 & 0.844 & 0.828 & 0.843 & 0.763 & 0.884 & 0.867 & 0.814 & 0.879 & 0.865 & 0.790     \\
         Improvement             & 0.030 & 0.050 & 0.046 & 0.007 & 0.007 & 0.006 & 0.067 & 0.023 & 0.056 & 0.022 & 0.032 & 0.024 & 0.013 & 0.017 & 0.024     \\
    \bottomrule
    \end{tabular}
    \begin{flushleft}
    BicliqueEncoder-NC stands for training BicliqueEncoder without formal concepts training samples.
    \end{flushleft}
    \label{ablation}
\end{sidewaystable}

The results indicate that learning information from formal concepts improves the prediction performance across these datasets, although the extent of improvement varies. 
% Particularly, the improvement on BMS-POS is minimal compared to the other datasets. This may be due to the relatively small number of the extracted concepts used to train the model in comparison to the other datasets. This demonstrates that extracting more concepts and utilizing them to train the model can enhance the prediction performance. 
The varying importance of the information contained in the extracted concepts for link prediction may contribute to the differences in improvement across datasets. In some datasets, the extracted concepts might be more representative of the entire concept lattice and thus more useful for improving link predictions compared to other datasets. Another possible reason is that the maximal length of concepts varies across datasets, resulting in different input lengths for the model. Longer input sequences may make it harder to train the model, which counterbalances the improvement brought by adding concepts as training samples to some extent.

The results also demonstrate that our proposed graph Transformer network possesses the capability of learning useful information for link prediction from raw bipartite networks. By solely using formal contexts as training samples, our model can directly learn from bipartite networks. Other baseline methods, on the other hand, directly take bipartite networks as input. From the results, we can find that with the same bipartite network as input, our model still has better prediction performance than other methods on the CTD and Bonanza datasets and comparable performance with SBGNN on the other three datasets. This suggests that our proposed graph transformer network can effectively extract useful information from bipartite networks for link prediction.

 \subsection{Supplementary Experiment}\label{subsec4}
 
To evaluate our method on small datasets and compare it with the previous FCA-based methods, we apply it to the datasets used in the publication where BERT4FCA was proposed~(\cite{bert4fca}). We apply BicliqueEncoder on these datasets and compare the prediction performance and execution time with BERT4FCA. We choose not to further compare our method with other FCA-based methods on the above five datasets because first, BERT4FCA was proven the best FCA-based link prediction method~(\cite{bert4fca}); second, previous methods are extremely time-consuming and require significant computational resources that is unavailable to us. Since the sizes of these small datasets are manageable, we extract all the concepts and utilize them to make link predictions in BicliqueEncoder. We depict the features of these datasets in~\autoref{description1}. Please refer to~(\cite{bert4fca}) for a detailed description of each dataset. The prediction results are reported in~\autoref{supplementary} and the time consumption is reported in~\autoref{time}.

Notice that we only measure the execution time of the prediction methods, which excludes the execution time of the FCA process because this time is the same for the two methods. When measuring the time consumption of BicliqueEncoder, we set the number of epochs to 180 and the batch size to 24 for all five datasets. When measuring the time consumption of BERT4FCA, we set the batch size to 24 for all five datasets. The number of epochs during the pre-training step and fine-tune step for BMS-POS-small and iJO1366 datasets is set to 90, and the number of epochs during the pre-training step and fine-tune step for ICFCA, Kyeword-Paper and Review datasets is set to 180. The reason for different training epochs is that it takes too long for both BMS-POS-small and iJO1366 datasets to train the BERT4FCA model. 

\begin{table}[!htbp]
\begin{adjustwidth}{-2.25in}{0in}
    \centering
    \caption{\bf The features of the five small datasets.}
    \begin{tabular}{ccccccc}
    \toprule
         Dataset          &  Objects&  Attributes&  Incidences&  Concepts\\
    \midrule
         ICFCA            & 351 & 12614 & 14445 & 878 \\

         BMS-POS-small    & 468 & 1946 & 7376 & 7791  \\

         Keyword-Paper    & 162 & 5206 & 7648 & 1713 \\

         Review           & 181 & 304 & 465 & 281  \\

         iJO1366          & 1805 & 2583 & 10183 & 5595  \\
                                      
    \bottomrule
    \end{tabular}
    \begin{flushleft}
    BMS-POS-small dataset is extracted from the full BMS-POS dataset.
    \end{flushleft}
    \label{description1}
\end{adjustwidth}
\end{table}

\begin{sidewaystable}[!htbp]
% \begin{adjustwidth}{-2.75in}{0in}
    \small
    \centering
    \caption{\bf The results for the supplementary experiment.}
    \begin{tabular}{c|p{0.8cm}p{0.8cm}p{0.8cm}|p{0.8cm}p{0.8cm}p{0.8cm}|p{0.8cm}p{0.8cm}p{0.8cm}|p{0.8cm}p{0.8cm}p{0.8cm}|p{0.8cm}p{0.8cm}p{0.8cm}}
   % \begin{tabular}{c|cccc|cccc|cccc|cccc|cccc}
    % \hline
    \toprule
         \multirow{2}{*}{Method} & \multicolumn{3}{c|}{ICFCA} & \multicolumn{3}{c|}{BMS-POS-small} & \multicolumn{3}{c|}{Keyword-Paper} & \multicolumn{3}{c|}{Review} & \multicolumn{3}{c}{iJO1366} \\
         \cmidrule{2-16}
         & $F_1$ & AUC & AUPR &  $F_1$ & AUC & AUPR  & $F_1$ & AUC & AUPR  & $F_1$ & AUC & AUPR & $F_1$ & AUC & AUPR \\
    \midrule
         BERT4FCA                & \textbf{0.741} & \textbf{0.812} & \textbf{0.777} & 0.725 & \textbf{0.823} & \textbf{0.788} & \textbf{0.744} & 0.765 & \textbf{0.711} & \textbf{0.677} & \textbf{0.675} & \textbf{0.690} & \textbf{0.701} & \textbf{0.744} & \textbf{0.761}   \\
         BicliqueEncoder         & 0.740 & 0.809 & 0.776  & \textbf{0.748} & 0.816 & 0.779 & 0.742 & \textbf{0.769} & \textbf{0.711} & 0.675 & 0.669 & 0.679 & 0.699 & 0.741 & 0.757   \\
         BERT4FCA-NC             & 0.723 & 0.801 & 0.766  & 0.701 & 0.800 & 0.760 & 0.724 & 0.748 & 0.707 & 0.667 & 0.660 & 0.629 & 0.695 & 0.740 & 0.749 \\
         BicliqueEncoder-NC      & 0.737 & 0.803 & 0.771 & 0.748 & 0.815 & 0.773 & 0.731 & 0.748 & 0.709 & 0.670 & 0.660 & 0.656 & 0.694 & 0.740 & 0.750   \\
    % \hline
    \bottomrule
    \end{tabular}
    \begin{flushleft}
    -NC stands for training models without formal concepts.
    \end{flushleft}
    \label{supplementary}
% \end{adjustwidth}
\end{sidewaystable}

\begin{table}[!htbp]
\begin{adjustwidth}{-2.75in}{0in}
    \small
    \centering
    \caption{\bf The results for the supplementary experiment.}
   \begin{tabular}{c|r|r|r|r|r}
   % \begin{tabular}{c|p{0.8cm}|p{0.8cm}|p{0.8cm|p{0.8cm}|p{0.8cm}}
    \toprule
         % \multirow{1}{*}{Method} & \multicolumn{1}{c|}{ICFCA} & \multicolumn{1}{c|}{BMS-POS-small} & \multicolumn{1}{c|}{Keyword-Paper} & \multicolumn{1}{c|}{Review} & \multicolumn{1}{c}{iJO1366} \\
         Method & ICFCA & BMS-POS-small & Keyword-Paper & Review & iJO1366 \\
    \midrule
         BERT4FCA               &  12149s &  73621s &  9957s &  373s &  44099s   \\
         BERT4FCA-NC            &   2339s &  11704s &  1910s &  180s &   5824s   \\
         BicliqueEncoder        &   6647s &  13448s &  5222s &  296s &   9057s   \\
         BicliqueEncoder-NC     &   1866s &  11319s &  1624s &   47s &   3960s   \\ 
    % \hline
    \botrule
    \end{tabular}
    \begin{flushleft}
    -NC stands for training models without concepts; ``s'' means seconds.
    \end{flushleft}
    \label{time}
\end{adjustwidth}
\end{table}

From the prediction results, we can observe that BERT4FCA has slightly better prediction performance than BicliqueEncoder across these five small datasets. We analysis it because BERT4FCA can learn additional information from concept lattices during the pre-training step and utilize it for link prediction. 

However, from the time consumption results, we can find that BERT4FCA requires significantly more execution time compared to BicliqueEncoder. While learning more information from concept lattices can lead to improved prediction performance, it comes at the cost of significantly increased computational resources. In BicliqueEncoder, we strike a balance between execution time and prediction performance to ensure applicability to large datasets while maintaining acceptable performance levels.

By comparing the results between BERT4FCA without concepts as training samples (which is denoted as BERT4FCA-NC) and BicliqueEncoder without concepts as training samples (which is denoted as BicliqueEncoder-NC), we can find that BicliqueEncoder-NC can achieve a better prediction performance with a lower time consumption. This indicates that our proposed model is more effective and efficient than the model used in BERT4FCA on capturing useful information from raw bipartite networks to make link predictions.

\subsection{Experimental environment and parameters}\label{subsec5}
The experiments are conducted on a Windows 11 server with 64GB RAM, an AMD Ryzen 9 7900X CPU, and an NVIDIA GeForce RTX 4090 GPU.

The codes for the deep-learning part of our method are implemented with Python 3.8.18 and Pytorch 2.1.0. The code for the algorithm for extracting formal concepts is written in C++ and compiled by MinGW 13.2.0.

The dimension of input embeddings of the Transformer encoder is 768, the dimension of the hidden layer in the Transformer encoder is 3072. The dimension of the hidden layer in MLP is 512. 

\section{Conclusion and Future Works}\label{sec6}

In this paper, we proposed BicliqueEncoder, a practical FCA-based method for link prediction in bipartite networks using the Transformer encoder. Our method extends the applicability of FCA-based link prediction approaches to large datasets through the utilization of iceberg concept lattices. We believe this strategy could be adapted to enhance the scalability of other FCA-based methods. Experimental results indicate that our method outperforms baseline methods such as heuristic node similarity-based methods, matrix factorization (MF), random walk node embedding method, and SBGNN. Supplementary experiments demonstrate that our method can achieve slightly lower prediction performance with a significantly shorter execution time compared to the previous FCA-based bipartite link prediction method, BERT4FCA, on small datasets. Moreover, the ablation experiment further reveals that learning information from concepts contributes to prediction performance, and our proposed model can efficiently and effectively extract useful information from bipartite networks for link prediction.

In future work, we plan to propose a method to determine the optimal number and length of extracted concepts without relying on prior experience or costly trials, thereby improving the practicality of our method. We also intend to explore alternative approaches for extracting significant formal concepts from concept lattices, beyond the use of iceberg concept lattices, to enhance the efficiency of FCA-based methods for bipartite link prediction. Additionally, we aim to extend our method to non-bipartite and heterogeneous networks to broaden its applicability. We also plan to develop a method that enables the model to learn the order relations of concepts within concept lattices. These order relations, as demonstrated in the BERT4FCA research, are useful for bipartite link prediction. We expect the prediction performance can be further improved by learning this information. 

% \bibliography{sn-bibliography}% common bib file
% if required, the content of .bbl file can be included here once bbl is generated
%\input sn-article.bbl

\newpage
\begin{appendices}
\section{The Pseudocode for Significant Concept Extraction}\label{LCM}

In the ``Significant Concept Extraction'' step, we set the upper and lower bounds on the sizes of extents and intents and only extract concepts in which the sizes of extents and intents fall in this range. Such a searching process is achieved by an algorithm derived from the \textit{LCM} algorithm~(\cite{LCM}). The pseudocode of the searching algorithm is presented in~\autoref{LCM_modify}.

\begin{algorithm}[H]
\caption{The LCM algorithm for extracting significant concepts.}
\label{LCM_modify}
\hspace*{\algorithmicindent} \textbf{Input} A formal context $\mathbb{K}=(G,M,I)$; the upper bound for the sizes of extents $u_1$; the lower bound for the sizes of the extents $l_1$; the upper bound for the sizes of intents $u_2$; the lower bound for the sizes of intents $l_2$.\\
\hspace*{\algorithmicindent} \textbf{Output} A list of significant formal concepts extracted from $\mathbb{K}$.
\begin{algorithmic}[1]
\Procedure{Pref}{$b$}
  \State{Return $\bigcup_{1\leq i\leq b}m_b$}
\EndProcedure
\Procedure{Closure}{$B$}
  \State{$A_1\leftarrow \{g\in G\ |\ \forall m\in B,\ (g,m)\in I\}$}
  \State{$B_1\leftarrow \{m\in M\ |\ \forall g\in A_1,\ (g,m)\in I\}$}
  \State{Return $(A_1,B_1)$}
\EndProcedure
\Procedure{LCM}{$(A,B,u_1,u_2,l_1,l_2)$}
  \If {$|A| \leq u_1$ and $|B| \geq l_2$}
    \State {Output $(A,B)$}
  \EndIf
  \For {$b\leftarrow |M|\ \mathrm{downto}\ 1$}
    \State {$(A_1,B_1)\leftarrow \textsc{Closure}(B\cup {m_b})$}
    \If {$(B_1 - B)\cap \textsc{Pref}(b - 1)=\emptyset$ and $|A_1| \geq l_1$ and $|B_1| \leq u_2$}
        \State {Call \textsc{LCM}$((A_1,B_1))$}
    \EndIf
  \EndFor
\EndProcedure
\State{Call \textsc{LCM}$(\textsc{Closure}(\emptyset))$}
\end{algorithmic}
\label{basicalgo}
\end{algorithm}

\section{The details of data preparation}\label{data preparation}
In this section, we provide a detailed explanation of the ``Training Data Preparation''. The training samples consist of concept samples, generated from the extracted concepts, and context samples, derived from the original formal context. The process involves generating four intermediate sets, followed by the generation of concept samples and context samples, and finally, padding the samples. We will explain each step in detail and provide the corresponding pseudocode. 

\subsection{Intermediate sets}\label{Appdedix B1}
First we need to generate four intermediate sets. The extents are collected into a set $E_\mathrm{p}$ and all intents are collected into a set $I_\mathrm{p}$. Then, we are to generate two distractor sets $E_\mathrm{n}$ and $I_\mathrm{n}$, representing non-extent object sets and non-intent attribute sets, correspondingly. First, for all extracted significant concepts, we split them into extents and intents. The extents are collected into a set $E_\mathrm{p}$ and all intents are collected into a set $I_\mathrm{p}$. 

Then, we are to generate two distractor sets $E_\mathrm{n}$ and $I_\mathrm{n}$, representing non-extent object sets and non-intent attribute sets, correspondingly. For each extent $e_\mathrm{p} \in E_\mathrm{p}$, a distractor object set $e_\mathrm{n}$ is generated by randomly selecting $k$ percentage of objects from $e_p$ and replacing them with randomly selected objects from the entire object set. If we find this generated $e_n$ satisfies that $e_\mathrm{n}\notin E_\mathrm{p}$ and $e_\mathrm{n}\notin E_\mathrm{n}$, then we add it into $E_\mathrm{n}$ or otherwise we regenerate another $e_\mathrm{n}$. Similarly, for each intent $i_\mathrm{p} \in I_\mathrm{p}$, a distractor attribute set $i_\mathrm{n}$ is generated by randomly selecting $k$ attributes from $i_\mathrm{p}$ and replacing them with randomly selected attributes from the entire attribute set. If $i_\mathrm{p} \notin I_\mathrm{p}$ and $i_\mathrm{p} \notin I_\mathrm{n}$, it is added to $I_\mathrm{n}$. The pseudocode of the algorithm is presented in~\autoref{Intermediate Sets}.

\begin{algorithm}[H]
\caption{Generate Intermediate Sets}
\label{Intermediate Sets}
\hspace*{\algorithmicindent} \textbf{Input} The set of significant concepts $S$, the object set $G$, the attribute set $M$, the hyperparameter $k$ satisfying $0 < k < 1$. \\
\hspace*{\algorithmicindent} \textbf{Output} The extent set $E_\mathrm{p}$, the intent set $I_\mathrm{p}$, the distractor set for extents $E_\mathrm{n}$, the distractor set for intents $I_\mathrm{n}$, a mapping \textsc{Pos} that maps the distractor object/attribute set to its corresponding extent/intent.
\begin{algorithmic}[1]
\State Initialize $E_\mathrm{p}, I_\mathrm{p}, E_\mathrm{n}, I_\mathrm{n} \gets \emptyset$

\ForAll{$(A, B)\in S$} \Comment{\textit{Generate extents and intents}}
    \State $E_\mathrm{p} \gets E_\mathrm{p} \cup \{A\}$
    \State $I_\mathrm{p} \gets I_\mathrm{p} \cup \{B\}$
\EndFor

\ForAll{$A=\{a_1,a_2,\ldots,a_{|A|}\} \in E_\mathrm{p}$} \Comment{\textit{Generate distractor object sets}}
    \Repeat
        \State{ $n_1\leftarrow \lfloor k\cdot |A| \rfloor$}
        \State{$A'\leftarrow A$}
        \State{Generate a random sequence $L\leftarrow \{l_1,l_2,l_3,\ldots,l_{n_1}\}$ such that $1\leq l_1 < l_2 < \cdots < l_{n_1}\leq |A|$}
        \For {$i\leftarrow 1$ to $n_1$}
            \State{Randomly choose $g\in G$ such that $g\neq a_{l_i}$}
            \State{$A'\leftarrow (A' - \{a_{l_i}\})\cup \{g\}$}
        \EndFor
    \Until{$A' \notin E_\mathrm{p}$ and $A' \notin E_\mathrm{n}$}
    \State $E_\mathrm{n} \gets E_\mathrm{n} \cup \{A'\}$
    \State{$\textsc{Pos}(A)\leftarrow A'$}
\EndFor

\ForAll{$B=\{b_1,b_2,\ldots,b_{|B|}\}\in I_\mathrm{p}$} \Comment{\textit{Generate distractor attribute sets}}
    \Repeat
        \State{ $n_2\leftarrow \lfloor k\cdot |B| \rfloor$}
        \State{$B'\leftarrow B$}
        \State{Generate a random sequence $L\leftarrow \{l_1,l_2,l_3,\ldots,l_{n_2}\}$ such that $1\leq l_1 < l_2 < \cdots < l_{n_2}\leq |B|$}
        \For {$i\leftarrow 1$ to $n_2$}
            \State{Randomly choose $m\in M$ such that $m\neq b_{l_i}$}
            \State{$B'\leftarrow (B' - \{b_{l_i}\})\cup \{g\}$}
        \EndFor
    \Until{$B' \notin I_\mathrm{p}$ and $B' \notin I_\mathrm{n}$}
    \State{ $I_\mathrm{n} \gets I_\mathrm{n} \cup \{B'\}$}
    \State{$\textsc{Pos}(B)\leftarrow B'$}
\EndFor

\State{Output $E_\mathrm{p}, I_\mathrm{p}, E_\mathrm{n}, I_\mathrm{n}$ and $\textsc{Pos}$}
\end{algorithmic}
\end{algorithm}

\newpage
\subsection{Concept samples}\label{Appdedix B2}

After generating the four intermediate sets, we can generate the concept samples. The negative samples are generated from $E_\mathrm{n}$ and $I_\mathrm{n}$, denoted as $C_\mathrm{n}$. First, for each pair of $(e_\mathrm{p}, i_\mathrm{p})$ such that $e_\mathrm{p}\in E_\mathrm{p}$ and $i_\mathrm{p}\in I_\mathrm{p}$, If we find that $e_\mathrm{p}$ is strongly related to $i_\mathrm{p}$, that is, $e_\mathrm{p}\times i_\mathrm{p}\subseteq Y$, we treat the pair $(e_\mathrm{p},i_\mathrm{p})$ as a positive concept sample and add it to $C_\mathrm{p}$. Then, for each pair $(e_\mathrm{n},i_\mathrm{n})$ such that $e_\mathrm{n}\in E_\mathrm{n}$ and $i_\mathrm{n}\in i_\mathrm{n}$, suppose that $e_\mathrm{p}$ is generated from $e_\mathrm{n}$ and $i_\mathrm{p}$ is generated from $i_\mathrm{n}$, if $e_\mathrm{p}$ is strongly related to $i_\mathrm{p}$ and $e_\mathrm{n}$ is \textbf{not} strongly related to $i_\mathrm{n}$, then we consider that the pair $(e_\mathrm{n}, i_\mathrm{n})$ is a good distractor and should be added into the negative concept sample set $C_\mathrm{n}$. It can be easily derived that we should have $\|C_\mathrm{p}\|=\|C_\mathrm{n}\|$. The pseudocode of the algorithm is presented in~\autoref{Concept Samples}.

\begin{algorithm}
\caption{Generate Concept Samples}
\label{Concept Samples}
\hspace*{\algorithmicindent} \textbf{Input} The extent set $E_\mathrm{p}$, the intent set $I_\mathrm{p}$, the distractor set for extents $E_\mathrm{n}$, the distractor set for intents $I_\mathrm{n}$, the mapping \textsc{Pos} that maps the distractor object/attribute set to its corresponding extent/intent, the set of binary relations between extents and intents $Y$ \\
\hspace*{\algorithmicindent} \textbf{Output} Positive concept sample set $C_\mathrm{p}$, negative concept sample set $C_\mathrm{n}$
\begin{algorithmic}[1]
\State Initialize $C_p, C_n \gets \emptyset$
\ForAll{$A\in E_\mathrm{p}$} \Comment{\textit{Generate Positive Concept Samples}}
    \ForAll{$B\in I_\mathrm{p}$}
        \If{$A \times B \subseteq Y$} \Comment{\textit{Check if $A$ and $B$ form part of a formal concept}}
            \State $C_\mathrm{p} \gets C_\mathrm{p} \cup \{(A, B)\}$
        \EndIf
    \EndFor
\EndFor

\ForAll{$A\in E_\mathrm{n}$} \Comment{\textit{Generate Positive Concept Samples}}
    \ForAll{$B\in I_\mathrm{n}$}
        \State $A' \gets \textsc{Pos}(A)$
        \State $B' \gets \textsc{Pos}(B)$
        \If{$A' \times B' \subseteq Y$ and $A \times B \not\subseteq Y$} 
            \State $C_\mathrm{n} \gets C_\mathrm{n} \cup \{(A', B')\}$
        \EndIf
    \EndFor
\EndFor

\State{Output $C_\mathrm{p}, C_\mathrm{n}$} \Comment{\textit{$|C_\mathrm{p}| = |C_\mathrm{n}|$ is assured}}
\end{algorithmic}
\end{algorithm}

\newpage
\subsection{Context samples}\label{Appdedix B3}
Besides the concept samples, we still need to generate the context samples because the concept samples may leave out some information of some minor attributes or objects. First, for every $g \in G$ and $m \in M$ such that $(g, m) \in I$, we add the pair $(\{g\}, \{m\})$ as a positive sample and add it to $T_p$. Then, we randomly select a pair of an object $g_1 \in G$ and an attribute $m_1 \in M$. If we find $(g_1,m_1) \notin I$ and $(g_1,m_1) \notin T_n$, we create a pair $(\{g_1\},\{m_1\})$ as a negative sample and add it into $T_n$. We repeat the aforementioned random sampling step until $\|T_\mathrm{p}\|=\|T_\mathrm{n}\|$. The pseudocode of the algorithm is presented in~\autoref{Context Samples}. 

\begin{algorithm}[H]
\caption{Generate Context Samples}
\label{Context Samples}
\hspace*{\algorithmicindent} \textbf{Input} The object set $G$, the attribute set $M$, and the incidence matrix $I$ \\
\hspace*{\algorithmicindent} \textbf{Output} The positive context sample set $T_\mathrm{p}$ and the negative context sample set $T_\mathrm{n}$
\begin{algorithmic}[1]

\State Initialize $T_\mathrm{p}, T_\mathrm{n} \gets \emptyset$

\ForAll{$g \in G$} \Comment{\textit{Generate positive context samples}}
    \ForAll{$m \in M$}
        \If{$(g, m) \in I$}
            \State $T_\mathrm{p} \gets T_\mathrm{p} \cup \{(\{g\}, \{m\})\}$
        \EndIf
    \EndFor
\EndFor

\While{$|T_\mathrm{p}| > |T_\mathrm{n}|$} \Comment{\textit{Generate negative context samples}}
    \State Randomly select $g_1 \in G$ and $m_1 \in M$
    \If{$(g_1, m_1) \notin I$ and $(\{g_1\}, \{m_1\}) \notin T_\mathrm{n}$}
        \State $T_\mathrm{n} \gets T_\mathrm{n} \cup \{(\{g_1\}, \{m_1\})\}$
    \EndIf
\EndWhile

\State{Output $T_\mathrm{p}, T_\mathrm{n}$}
\end{algorithmic}
\end{algorithm}

\newpage
\subsection{Pad samples}\label{Appdedix B4}
After obtaining all training samples, we need to pad them to the same length with special tokens ``[PAD]'' so that they can be fed into the network. Suppose that $(A_1,B_1)\in C_\mathrm{p}\cup C_\mathrm{n}$ is concept sample with the largest object set and $(A_2,B_2)\in C_\mathrm{p}\cup C_\mathrm{n}$ is the concept sample with the largest attribute set. Then, for each sample $(X,Y)\in$ $C_\mathrm{p}\cup C_\mathrm{n}\cup T_\mathrm{p}\cup T_\mathrm{n}$, we convert $X$ to a vector $X_1$ and pad $\|A_1\| - \|X\|$ [PAD]s at the rear of $X_1$, and convert $Y$ to a vector $Y_1$ and pad $\|B_1\| - \|Y\|$ [PAD]s at the rear of $Y_1$. The padded training sample $(X_1, Y_1)$ is added to the final training set $T_\mathrm{F}$, with its label (positive or negative) set to be the same of that of $(X, Y)$. After this step, we will obtain the the final training set $T_\mathrm{F}$, with its label (positive or negative) set. The pseudocode of the algorithm is presented in~\autoref{Pad}. 

\begin{algorithm}
\caption{Pad Training Samples}
\hspace*{\algorithmicindent} \textbf{Input} Positive concept sample set $C_\mathrm{p}$, negative concept sample set $C_\mathrm{n}$, positive context sample set $T_\mathrm{p}$, negative context sample set $T_\mathrm{n}$, special token ``[PAD]''
\hspace*{\algorithmicindent} \textbf{Output} Padded training set $T_F$
\label{Pad}
\begin{algorithmic}[1]

\State Initialize $T_F \gets \emptyset$

\State{Initialize $l_1,l_2\leftarrow 0$}
\ForAll{$(A, B) \in C_\mathrm{p} \cup C_\mathrm{n}$} \Comment{\textit{Find the size of the largest object set}} 
    \State{$l_1\leftarrow \max\{l_1,|B|\}$}
\EndFor
\ForAll{$(A, B) \in C_\mathrm{p} \cup C_\mathrm{n}$} \Comment{\textit{Find the size of the largest attribute set}}
    \State{$l_2\leftarrow \max\{l_2,|B|\}$}
\EndFor

\ForAll{$(A, B) \in C_\mathrm{p} \cup C_\mathrm{n} \cup T_\mathrm{p} \cup T_\mathrm{n}$}
    \State $V_1\leftarrow (a_1,a_2,a_3,\ldots,a_{|A|},z_1,z_2,\ldots,z_{l_1-|A|})$ where $\bigcup_{1\leq i\leq |A|}\{a_i\}=A$ and $z_j=\mathrm{[PAD]}$ for all $1\leq j\leq l_1-|A|$
    \State $V_2\leftarrow (b_1,b_2,b_3,\ldots,b_{|B|},z_1,z_2,\ldots,z_{l_2-|B|})$ where $\bigcup_{1\leq i\leq |B|}\{b_i\}=B$ and $z_j=\mathrm{[PAD]}$ for all $1\leq j\leq l_2-|B|$
    
    \State $T_\mathrm{F}\leftarrow T_\mathrm{F}\cup (V_1, V_2)$ \Comment{\textit{The label of $(V_1,V_2)$ is set to be} TRUE \textit{if $(A, B)\in C_\mathrm{p}\cup T_\mathrm{p}$ or otherwise} FALSE}
\EndFor

\State{Output $T_F$}
\end{algorithmic}
\end{algorithm}

\end{appendices}

\end{document}